\documentclass[journal]{IEEEtran}
\usepackage{bm}
\usepackage{comment}
\usepackage{amsfonts} 
\usepackage{textcomp}
\usepackage{stmaryrd}
\usepackage{graphicx} 
\usepackage{verbatim}
\usepackage{multirow}
\usepackage{xspace}
\usepackage{amsmath}
\usepackage{hyperref}
\usepackage{subcaption}

\usepackage{booktabs}    
\usepackage{arydshln}    
\usepackage{wrapfig}
\usepackage{enumitem}
\setlist{nolistsep}

\newcommand{\eg}{\emph{e.g.}\xspace}
\newcommand{\ie}{\emph{i.e.}\xspace}
\newcommand{\etal}{\emph{et al.}\xspace}

\ifCLASSINFOpdf
\else
\fi
\begin{document}
%
\title{SupeRANSAC: One RANSAC to Rule Them All}
%
%
%

\author{Daniel Barath
\thanks{D. Barath was with ETH Zurich,
Switzerland, e-mail: dbarath@ethz.ch}
\thanks{Manuscript received April 19, 2005}}

%
%

\markboth{}%
{Barath \MakeLowercase{\textit{et al.}}: SupeRANSAC: One RANSAC to Rule Them All}
%



\maketitle

\begin{abstract}
    Robust estimation is a cornerstone in computer vision, particularly for tasks like Structure-from-Motion and Simultaneous Localization and Mapping. 
    RANSAC and its variants are the gold standard for estimating geometric models (\eg, homographies, relative/absolute poses) from outlier-contaminated data.  
    Despite RANSAC's apparent simplicity, achieving consistently high performance across different problems is challenging. 
    While recent research often focuses on improving specific RANSAC components (\eg, sampling, scoring), overall performance is frequently more influenced by the "bells and whistles" – the implementation details and problem-specific optimizations – within a given library.  
    Popular frameworks like OpenCV and PoseLib demonstrate varying performance, excelling in some tasks but lagging in others.  
    We introduce SupeRANSAC, a novel unified RANSAC pipeline, and provide a detailed analysis of the techniques that make RANSAC effective for specific vision tasks, including homography, fundamental/essential matrix, and absolute/rigid pose estimation.  
    SupeRANSAC is designed for consistent accuracy across these tasks, improving upon the best existing methods by, for example, 6 AUC points on average for fundamental matrix estimation.  We demonstrate significant performance improvements over the state-of-the-art on multiple problems and datasets.  Code: \url{https://github.com/danini/superansac}
\end{abstract}

\begin{IEEEkeywords}
RANSAC, robust estimation, relative pose, absolute pose, rigid pose
\end{IEEEkeywords}

%
\IEEEpeerreviewmaketitle

\section{Introduction}

Robust model estimation is fundamental in computer vision, crucial for tasks like visual localization~\cite{sarlin2019coarse}, Structure-from-Motion (SfM)~\cite{schonberger2016structure,pan2024global}, Simultaneous Localization and Mapping (SLAM)~\cite{mur2015orb,martinez2022ransac}, and object pose estimation~\cite{hodan2020epos,ornek2025foundpose}. 
The Random Sample Consensus (RANSAC) algorithm~\cite{fischler1981random} is the standard approach for robust estimation in the presence of noisy, outlier-contaminated data. 
RANSAC iteratively selects minimal data samples, estimates model parameters, and assesses model quality by counting inliers. 

Despite its popularity, RANSAC's sensitivity real-world noise distributions, to its parameters (\eg, the inlier-outlier threshold) and susceptibility to ill-conditioned samples have motivated extensive research~\cite{cvpr2020ransactutorial}.  
Improvements include novel sampling methods for early good model discovery~\cite{chum2005matching,torr2002napsac,cavalli2022nefsac,brachmann2019neural}, new scoring functions that better model noise distributions~\cite{torr2000mlesac,barath2021marginalizing,barath2020magsac++,barath2022learning}, refinement algorithms to mitigate the effects of noisy samples~\cite{chum2003locally,lebeda2012fixing,barath2022learning}, and degeneracy checks~\cite{chum2005two,cavalli2022nefsac}. Recent methods also explore learning specific RANSAC components (\eg, inlier probabilities, scoring/sampling)~\cite{brachmann2019neural,barath2022learning,cavalli2022nefsac} or enabling end-to-end differentiable estimation for training networks~\cite{brachmann2017dsac,wei2023generalized}.

Most publications proposing enhancements to individual RANSAC components demonstrate improved accuracy or efficiency, often attributing gains solely to the technique proposed in the current paper, while providing limited details on the rest of the pipeline.  Frameworks like USAC~\cite{raguram2012usac} and VSAC~\cite{ivashechkin2021vsac} emphasize that optimal performance requires careful attention to \textit{all} components.  However, this holistic view is often overlooked in recent literature.  Crucial components, such as minimal/non-minimal solver selection, efficient model/sample degeneracy checks, and advanced local optimization, can profoundly impact overall performance.  While USAC and VSAC offer valuable insights, their accuracy lags behind modern frameworks like PoseLib~\cite{poselib}.

Today, there are many robust estimation frameworks, including PoseLib~\cite{poselib}, pyCOLMAP~\cite{schonberger2016structure}, OpenCV~\cite{bradski2000opencv}, USAC~\cite{raguram2012usac}, VSAC~\cite{ivashechkin2021vsac}, GC-RANSAC~\cite{barath2018graph}, and MAGSAC++~\cite{barath2020magsac++}.  
Each framework possesses distinct strengths and weaknesses. 
For instance, PoseLib excels in absolute pose and essential matrix estimation but performs poorly on homographies and lacks code for rigid pose estimation for 3D point clouds.  Conversely, GC-RANSAC and MAGSAC++ are strong in homography, fundamental matrix, and rigid pose estimation but underperform in essential matrix and absolute pose estimation compared to PoseLib. These variations in performance stem from differing design choices and the inclusion of specific optimizations and checks -- the "bells and whistles" -- even when the core components (samplers, minimal solvers, local optimization) are similar.

\begin{figure}[t!]
    \centering
    \includegraphics[width=0.49\columnwidth]{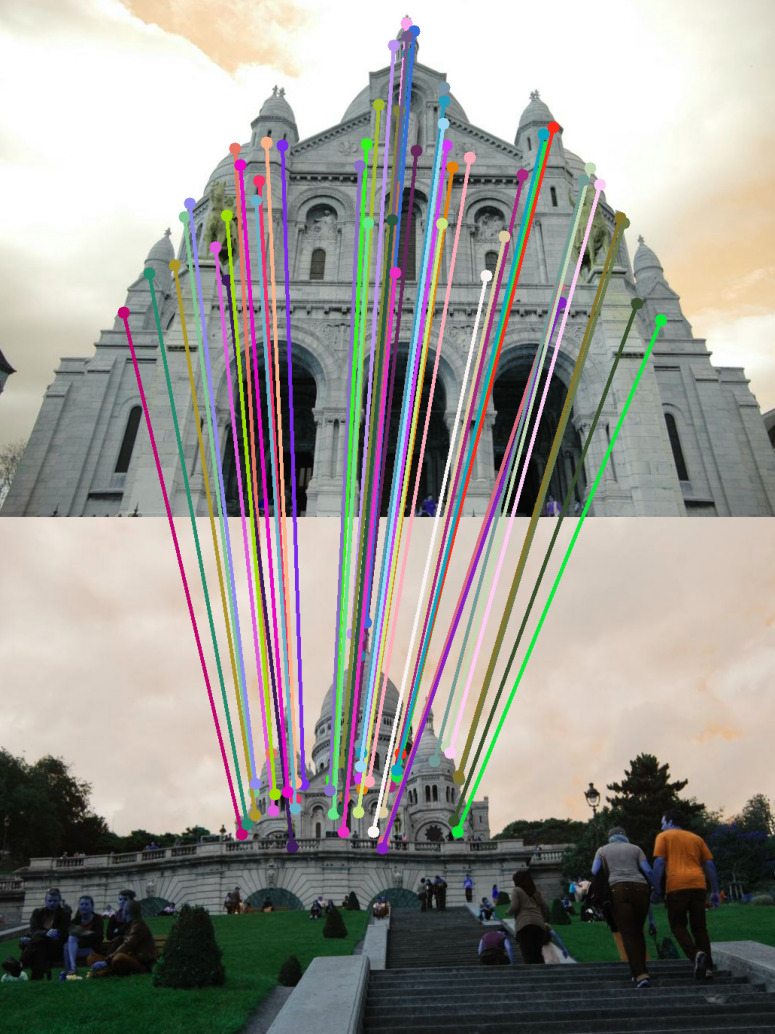}
    \includegraphics[width=0.49\columnwidth]{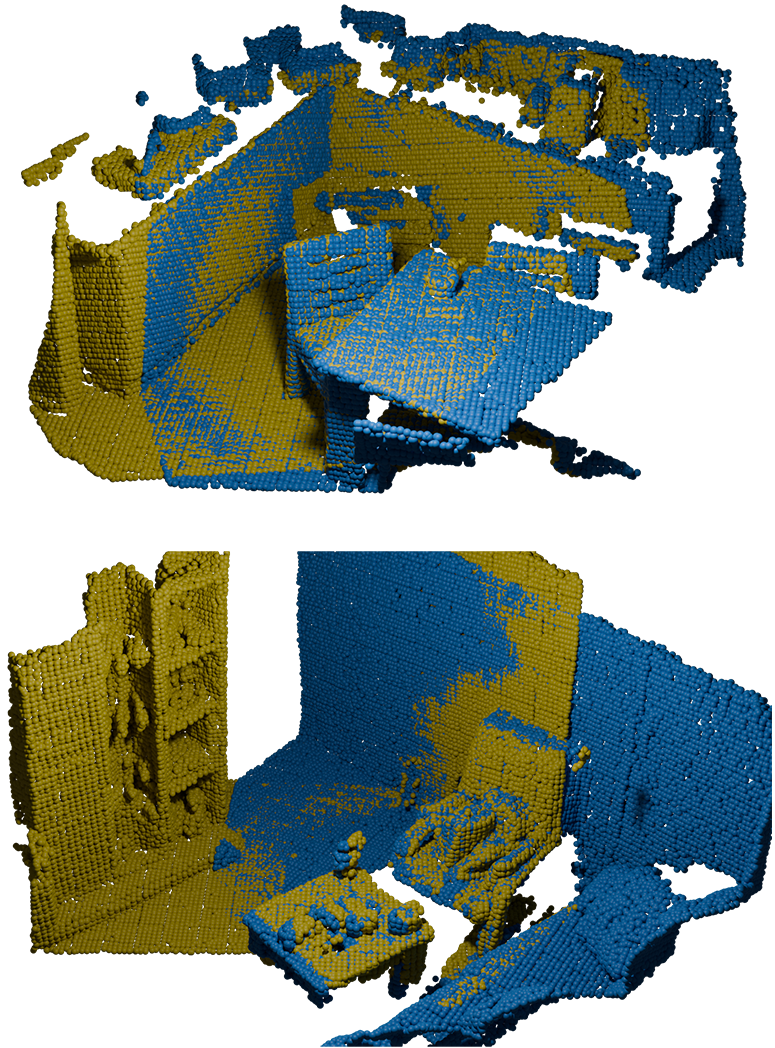}
    \caption{
    Inliers found by SupeRANSAC for fundamental matrix estimation (left plot; PhotoTourism dataset~\cite{snavely2006photo}) and 3D point cloud alignment (right; 3DMatch dataset~\cite{zeng20173dmatch}). SupeRANSAC achieves state-of-the-art results on 11 public large-scale datasets for homography, fundamental and essential matrix, rigid and absolute pose estimation.}
\end{figure}

This paper revisits robust estimation from a foundational perspective, presenting a novel framework, SupeRANSAC, that achieves consistently high accuracy across common vision problems: homography, essential and fundamental matrix estimation, and rigid and absolute pose estimation. We meticulously design a novel pipeline to ensure that all problem-specific details (\eg, the choice of minimal and non-minimal solvers) are aligned with the goal of maximizing accuracy.  We demonstrate the effectiveness of the proposed SupeRANSAC on 11 large-scale public datasets, using both state-of-the-art sparse~\cite{detone2018superpoint,lindenberger2023lightglue} and dense features~\cite{edstedt2024roma}.  
SupeRANSAC substantially outperforms existing public frameworks across the tested problems.

\section{Related Work}

Robust estimation, the process of fitting a model to data contaminated with outliers, is a cornerstone of computer vision. In this domain, each data point typically represents a constraint on the underlying model. The dominant paradigm for robust estimation relies on randomized sampling methods, particularly RANSAC (Random Sample Consensus)~\cite{fischler1981random}, a highly successful and widely adopted technique. Given RANSAC's foundational importance and its widespread application, extensive research has been dedicated to enhancing its core aspects: robustness to high outlier rates, accuracy of the estimated model, and computational efficiency. These efforts can be broadly categorized into advancements in sampling strategies, model scoring mechanisms, local optimization and verification techniques, and, more recently, learned outlier pruning.

\paragraph{Sampling Strategies}  
The efficiency of RANSAC heavily depends on the probability of drawing a minimal sample composed entirely of inliers. 
Consequently, numerous approaches aim to improve upon the standard uniform sampling strategy, thereby increasing the probability of discovering a good model early in the iterative process. 
PROSAC~\cite{chum2005matching} exemplifies quality-driven sampling by leveraging a prior ordering of putative correspondences, typically based on matching scores, to sample more promising points first before gradually transitioning to uniform sampling. 
Other strategies exploit spatial locality or structural cues, such as conditional sampling that selects neighboring correspondences~\cite{torr2002napsac,barath2019progressive,barath2023finding}, or methods that sample from pre-identified clusters~\cite{ni2009groupsac}, assuming inliers exhibit such groupings. 
Adaptive sampling methods, explored in works like~\cite{fan2009adaptive,piedade2023bansac,wei2023adaptive}, dynamically adjust the sampling process, often using heuristics derived from the success or failure of previous samples to bias subsequent selections toward more promising regions of the data space.

A significant recent trend involves leveraging machine learning to guide the sampling process. Neural-Guided RANSAC~\cite{brachmann2019neural}, for instance, learns a categorical sampling distribution over points using an approximate expected gradient of the RANSAC objective. More recent works, such as $\nabla$-RANSAC~\cite{wei2023generalized}, have advanced this direction by proposing fully differentiable RANSAC pipelines. 
$\nabla$-RANSAC learns a categorical sampling distribution via the Gumbel reparameterization trick, enabling end-to-end training and facilitating a tighter integration of data-driven priors into the sampling mechanism. The overarching goal of these learning-based sampling strategies is to significantly increase the probability of drawing outlier-free minimal samples by prioritizing points that are more likely to be inliers based on learned patterns.

\paragraph{Model Scoring}  
Beyond the rudimentary inlier counting used in vanilla RANSAC, several work have proposed more sophisticated model scoring functions. 
The motivation is twofold: to improve the quality of the selected model and to reduce sensitivity to the critical inlier-outlier threshold. 
Probabilistic approaches, such as Bayesian RANSAC~\cite{torr2002bayesian} and MLESAC~\cite{torr2000mlesac}, employ Bayesian inference and maximum likelihood estimation, respectively, for a more principled estimation of model parameters and inlier sets. 
Other methods, like MINPRAN~\cite{stewart1995minpran} and a contrario RANSAC~\cite{moisan2012automatic}, focus on automatically determining an optimal, data-dependent threshold for each minimal sample model by minimizing the probability of the observed point distributions being a result of random chance. 
DSAC~\cite{brachmann2017dsac} introduced a differentiable scoring mechanism based on probabilistic model selection, facilitating the joint learning of model scoring and the sampling distribution. 
More recently, MAGSAC++~\cite{barath2020magsac++,barath2021marginalizing} proposed to marginalize over a range of noise scales (and thus, implicitly, over multiple thresholds), significantly improving robustness to manual threshold selection. 
Barath \etal~\cite{barath2022learning} proposed learning the entire scoring function in a data-driven manner, using point-to-model residual histograms as input, thereby aiming to eliminate the need for a manually defined threshold.

\paragraph{Local Optimization and Degeneracy Checks}  
To refine the accuracy of models estimated from minimal samples, many RANSAC variants incorporate local optimization (LO), degeneracy checks, and preemptive verification steps. 
LO-RANSAC~\cite{chum2003locally} was a seminal work demonstrating that performing local optimization on the parameters of the best models discovered during the consensus-seeking search can significantly improve model accuracy. 
This typically involves re-estimating the model using all its consensus inliers. 
Subsequent research has further refined LO techniques~\cite{lebeda2012fixing} and integrated them with more powerful inlier/outlier segmentation methods, such as graph-cut based optimization~\cite{barath2018graph}, which can lead to more robust model estimates. 

To accelerate the RANSAC process, especially when dealing with large datasets or high outlier ratios, preemptive model verification techniques have been developed. 
The Sequential Probability Ratio Test (SPRT)~\cite{chum2008optimal} and, more recently, Space Partitioning RANSAC~\cite{barath2022space}, introduce methods to efficiently interrupt the score calculation for hypotheses that are statistically unlikely to surpass the quality of the current best model. Addressing another critical failure mode, DEGENSAC~\cite{chum2005two} specifically tackles the issue of degenerate minimal samples (checking for planar degeneracy in 2D-2D correspondences) that can lead to models with erroneously high consensus. QDEGSAC~\cite{frahm2006ransac} extends this by addressing quasi-degenerate models, actively seeking missing geometric constraints within the set initially labeled as outliers. Comprehensive frameworks like USAC~\cite{raguram2012usac} and VSAC~\cite{ivashechkin2021vsac} underscore the importance of a holistic pipeline design by integrating various state-of-the-art improvements, including adaptive sampling strategies, robust local optimization, and efficient verification tests, to achieve consistently high performance.

\paragraph{Learned Outlier Pruning}  
Distinct from methods that integrate learning into the core RANSAC loop (sampling or scoring), another line of research focuses on pre-processing the input data to prune putative outliers before RANSAC is even applied. 
These approaches aim to reduce the effective outlier ratio, thereby simplifying the task for the subsequent robust estimator. 
Early methods in this category often relied on handcrafted rules or statistical tests on local neighborhoods, such as GMS~\cite{bian2017gms} which verifies correspondence consistency using statistics of neighboring matches, or~\cite{cavalli2020handcrafted} which uses local affine model fitting for validation. 

More recently, deep learning techniques have become prevalent for this task. For example, Yi \etal~\cite{yi2018learning} utilized PointNet~\cite{qi2017pointnet} with batch normalization as a form of context normalization for learning to reject outliers in correspondence sets. Sun \etal~\cite{sun2020acne} introduced an attention mechanism to enhance the robustness of learned feature matching and outlier rejection. OANet~\cite{zhang2019learning} proposed an order-aware pooling mechanism for point features to learn soft clustering of inliers and outliers. Architectures like CLNet~\cite{zhao2021progressive} offer further improvements for progressive outlier pruning. MS2DG-Net~\cite{dai2022ms2dg} constructs a dynamic sparse semantic graph to better exploit spatial correlations among inliers for improved pruning. Some of these outlier filtering methods, such as~\cite{zhao2021progressive,dai2022ms2dg}, also internally generate model estimates or leverage consensus mechanisms to guide the outlier rejection process. Nevertheless, as highlighted in comprehensive studies and tutorials~\cite{cvpr2020ransactutorial}, applying a robust estimator like RANSAC to the filtered data consistently yields benefits in terms of final model accuracy and robustness.

While such learning-based pre-processing approaches offer advantages in many scenarios, we intentionally exclude them from our proposed pipeline in this work to preserve the out-of-the-box usability and general applicability of our method without reliance on prior training data or learned models.

\section{SupeRANSAC}\label{sec:superansac}

\begin{figure}[t!]
    \centering
    \includegraphics[width=1.0\columnwidth]{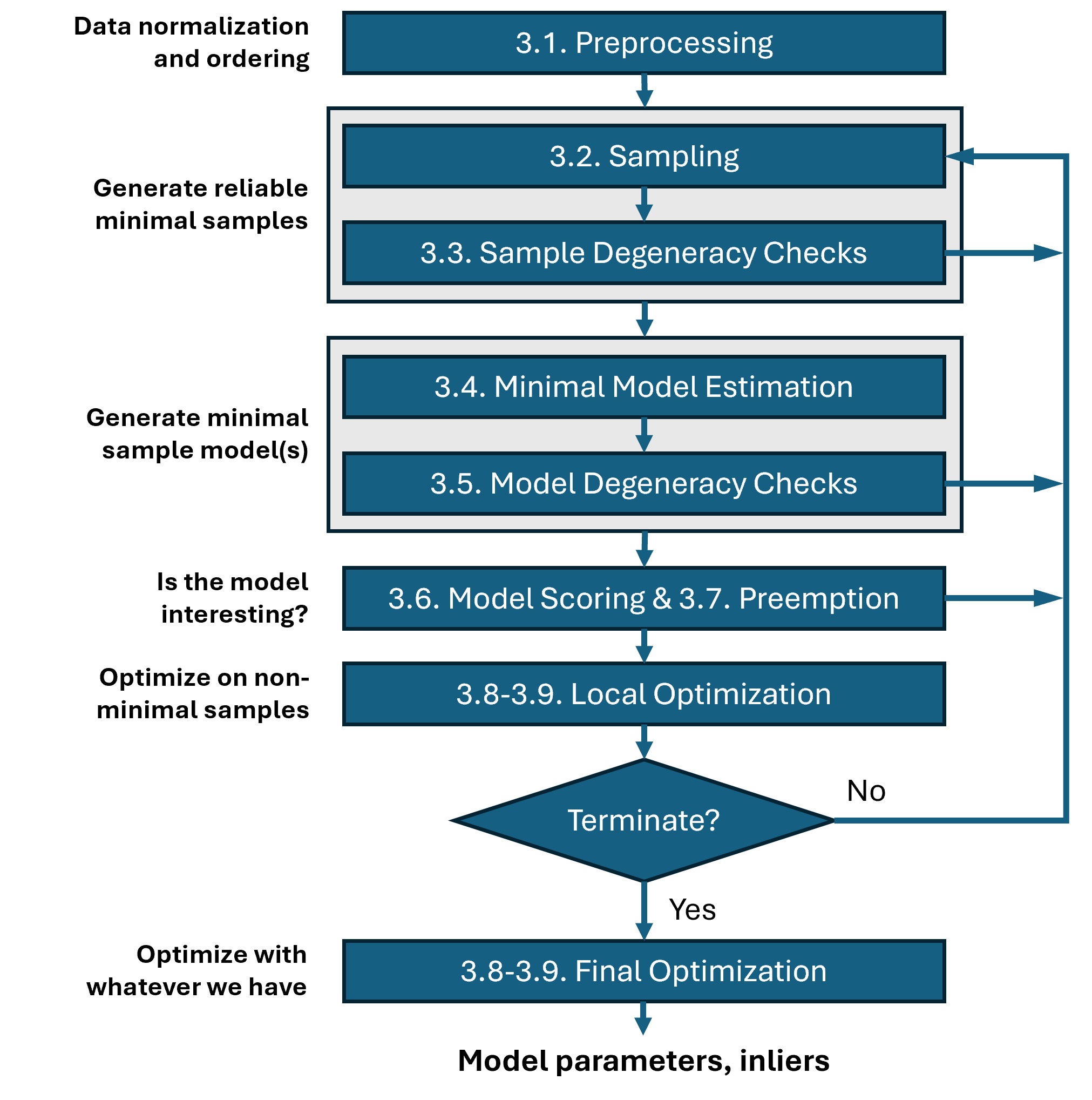}
    \caption{\textbf{Overview} of the SupeRANSAC framework, where each processing stage is specialized for the given geometric estimation problem (e.g., homography or fundamental matrix estimation). All components are detailed in Sec.~\ref{sec:superansac}.}
    \label{fig:pipeline}
\end{figure}

This section details each component of the proposed SupeRANSAC framework, visualized in Fig.~\ref{fig:pipeline}.   
While the pipeline is unified, each different estimation problem may require certain components to be adapted to account for the different mathematical constraints and characteristics they impose. 

\subsection{Preprocessing}

Effective data normalization is an essential preliminary step in robust estimation.
For two-view geometric problems where camera intrinsics are unknown (\eg, homography and fundamental matrix estimation), input correspondences are normalized such that their centroid is translated to the origin and their average distance to the origin is $\sqrt{2}$, following the seminal work of Hartley~\cite{hartley1997defense}. When camera intrinsic parameters are known (\eg, for essential matrix estimation or absolute pose estimation from 3D-to-2D correspondences), we normalize 2D image points by transforming them into normalized camera coordinates using the provided focal lengths and principal points. For problems involving 3D point clouds (\eg, rigid transformation estimation), we translate the points so that their centroid is at the origin; no scaling is applied in this case to preserve the original scene scale. Crucially, the inlier-outlier threshold $\tau$ is appropriately scaled in accordance with the applied normalization to maintain its geometric meaning.

\subsection{Sampling} 

In the RANSAC framework, the primary objective of the sampling stage is to efficiently identify a minimal subset of data points consisting entirely of inliers. Although improved sampling is often associated with faster convergence, its impact on the \textit{accuracy} of the final model is also profound, particularly in challenging scenarios characterized by low inlier ratios where exhaustive evaluation of all possible minimal sets is computationally prohibitive. Standard RANSAC employs a uniform random sampling strategy, treating all data points as equally likely to be part of an uncontaminated sample.

Drawing inspiration from the extensive literature on advanced sampling techniques~\cite{torr2002napsac,chum2005matching,brachmann2019neural,barath2019progressive,wei2023adaptive,piedade2023bansac,barath2023finding}, SupeRANSAC strategically incorporates two proven methods: \textit{PROSAC} (Progressive Sample Consensus)~\cite{chum2005matching} and \textit{P-NAPSAC} (Progressive NAPSAC)~\cite{barath2019progressive}.
PROSAC leverages prior information about the likelihood of each point being an inlier, typically derived from feature matching scores. It prioritizes sampling from the highest-quality points first, gradually expanding the sampling pool to include lower-quality data as iterations proceed. This strategy aims to find an all-inlier sample much faster than uniform sampling if the quality scores are informative.
P-NAPSAC extends PROSAC by integrating a spatial coherency prior. The initial point of a minimal sample is selected according to the PROSAC scheme, and subsequent points are preferentially drawn from a spatially localized region around this initial point, often defined by a gradually expanding hypersphere. For computational efficiency, our SupeRANSAC implementation approximates this localized sampling by employing multiple overlapping uniform grids and applying the PROSAC sampling logic locally within the grid cell containing the initially selected point.

The choice of sampling strategy in SupeRANSAC is adapted to the specific geometric estimation problem. P-NAPSAC is employed for homography estimation, rigid transformation fitting, and absolute pose estimation, where spatial proximity is often indicative of inlier relationships. However, for epipolar geometry estimation (fundamental or essential matrix), localized sampling can be detrimental, frequently leading to degenerate configurations (\eg, points concentrated on a small region or a plane) or yielding geometrically unstable models. Therefore, for epipolar geometry, SupeRANSAC defaults to the standard PROSAC sampling strategy to draw from a more globally distributed set of correspondences.

\subsection{Sample Degeneracy Checks} 

To accelerate the estimation process, SupeRANSAC incorporates sample degeneracy checks that assess the geometric viability of a minimal sample \emph{before} attempting model estimation. 
If a sample is deemed degenerate, it is rejected, and a new sample is drawn, thereby avoiding wasted computation on ill-conditioned inputs. 
We implement specific geometric degeneracy tests for homography estimation, rigid transformation fitting, and absolute pose estimation, as detailed below.

\paragraph{Homography Estimation}
For a homography to be well-defined by four point correspondences $(p_i, p_i')$, $i \in \{1,2,3,4\}$, two primary conditions must be met for the points in each image: 
(1) no three points should be collinear, 
and (2) the quadrilateral formed by the four points (\eg, $p_1p_2p_3p_4$ in sequence) should not be "twisted" or self-intersecting (\ie, it must be a simple polygon). 

Our primary geometric check focuses on the "twisted" quadrilateral condition, sometimes referred to as a planarity or orientation consistency check. Assuming a consistent cyclic ordering of points (\eg, $p_1, p_2, p_3, p_4$), the quadrilateral should not self-intersect. A self-intersecting, or "bow-tie", quadrilateral (\eg, where edges $p_1p_2$ and $p_3p_4$ cross) yields an infeasible homography for many applications. This check is performed for the point sets in both images. If the configuration in either image is degenerate, the sample is rejected.
To test if two line segments, say $S_1 = AB$ and $S_2 = CD$, intersect, we verify that points $A$ and $B$ lie on opposite sides of the line defined by $CD$, and simultaneously, points $C$ and $D$ lie on opposite sides of the line defined by $AB$. This can be implemented using 2D cross-products to check orientations:
\begin{itemize}
    \item $\text{Orientation}(C,D,A)$ and $\text{Orientation}(C,D,B)$ must have opposite signs (or one is zero).
    \item $\text{Orientation}(A,B,C)$ and $\text{Orientation}(A,B,D)$ must have opposite signs (or one is zero).
\end{itemize}
The orientation of three ordered points $(P_1, P_2, P_3)$ can be computed as $(x_2 - x_1)(y_3 - y_1) - (y_2 - y_1)(x_3 - x_1)$. If both conditions are met, the segments intersect. For the quadrilateral $p_1p_2p_3p_4$, we check if non-adjacent edges (\eg, $p_1p_2$ and $p_3p_4$) intersect. Such an intersection indicates a "twisted" sample, which is then rejected.

\paragraph{Rigid Transformation and Absolute Pose Estimation}

For both rigid transformation (from 3D-3D correspondences) and absolute pose estimation (from 2D-3D correspondences, typically P$n$P algorithms), the minimal sample involves three 3D points. A common degeneracy arises if these three 3D points are collinear. We detect this by computing the cross product of two vectors formed by the points, \eg, $\vec{v}_1 = P_2 - P_1$ and $\vec{v}_2 = P_3 - P_1$. If the magnitude of $\vec{v}_1 \times \vec{v}_2$ is close to zero, the points are collinear, and the sample is rejected. For rigid transformation estimation, this collinearity check is applied to the corresponding triplets of points in both 3D coordinate systems. For absolute pose estimation, the check is applied to the set of three 3D world points.

\paragraph{Essential and Fundamental Matrix Estimation}
Identifying degenerate sample configurations for relative pose estimation (\ie, fundamental or essential matrix) based solely on geometric properties of the minimal point correspondences (\eg, 5 points for E-matrix, 7 or 8 for F-matrix) is significantly more complex than for homographies or pose from 3D points. Common degeneracies, such as the planar scene degeneracy for fundamental matrix estimation or specific critical surfaces for essential matrix estimation, are not easily captured by inspecting only the minimal sample without attempting a solution. While learning-based approaches to detect such degeneracies prior to solving have been proposed~\cite{cavalli2022nefsac}, we currently do not incorporate such methods to maintain SupeRANSAC as a general-purpose, training-free framework that operates effectively out-of-the-box across diverse problem instances.

\begin{figure*}[t!]
    \centering
    \begin{subfigure}[t]{0.32\textwidth}
        \centering
        \includegraphics[width=1.0\columnwidth]{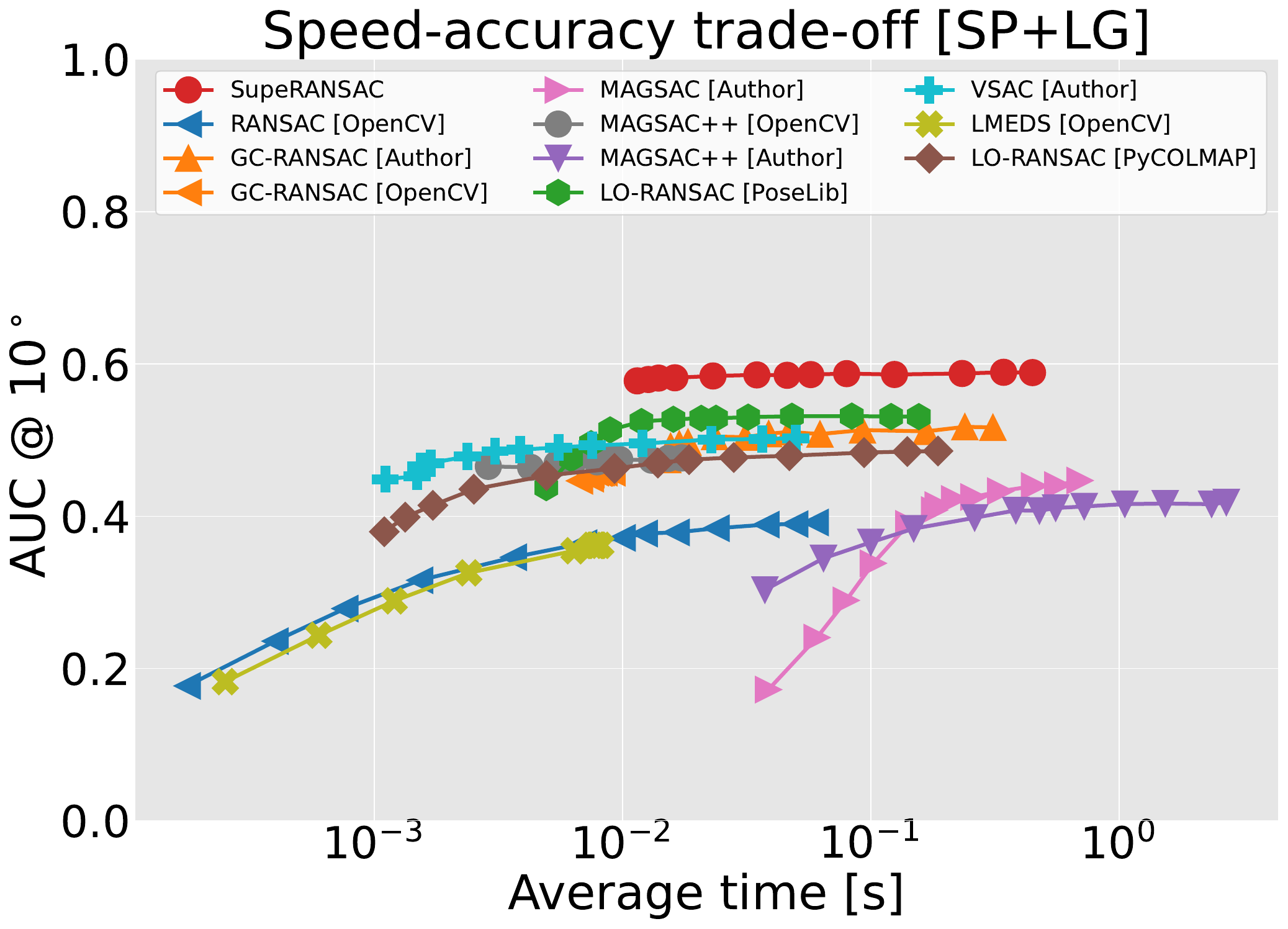}\\
        \includegraphics[width=1.0\columnwidth]{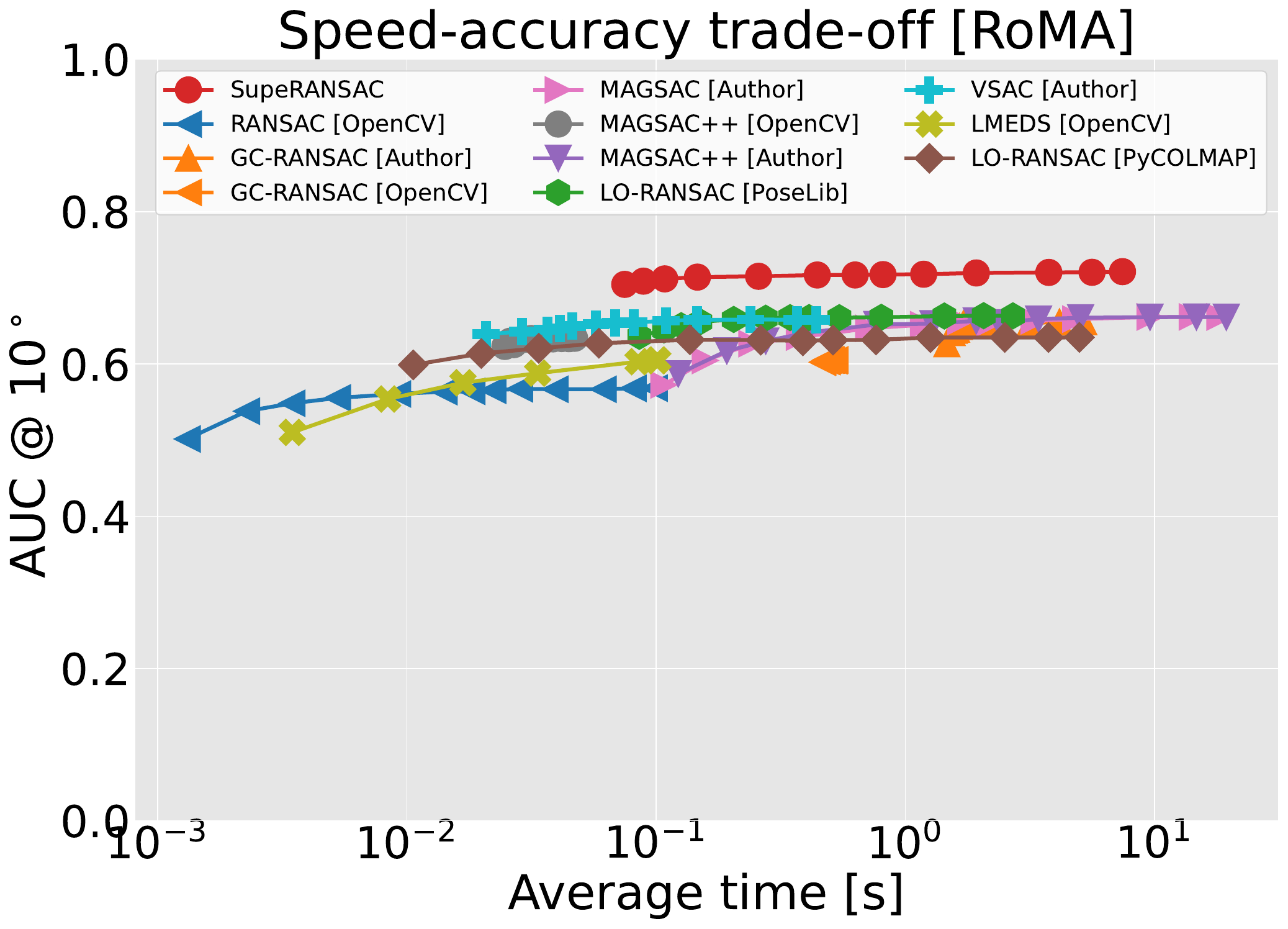}
        \caption{Fundamental matrix}
        \label{fig:accuracy_curves_F}
    \end{subfigure}%
    ~ 
    \begin{subfigure}[t]{0.32\textwidth}
        \centering
        \includegraphics[width=1.0\columnwidth]{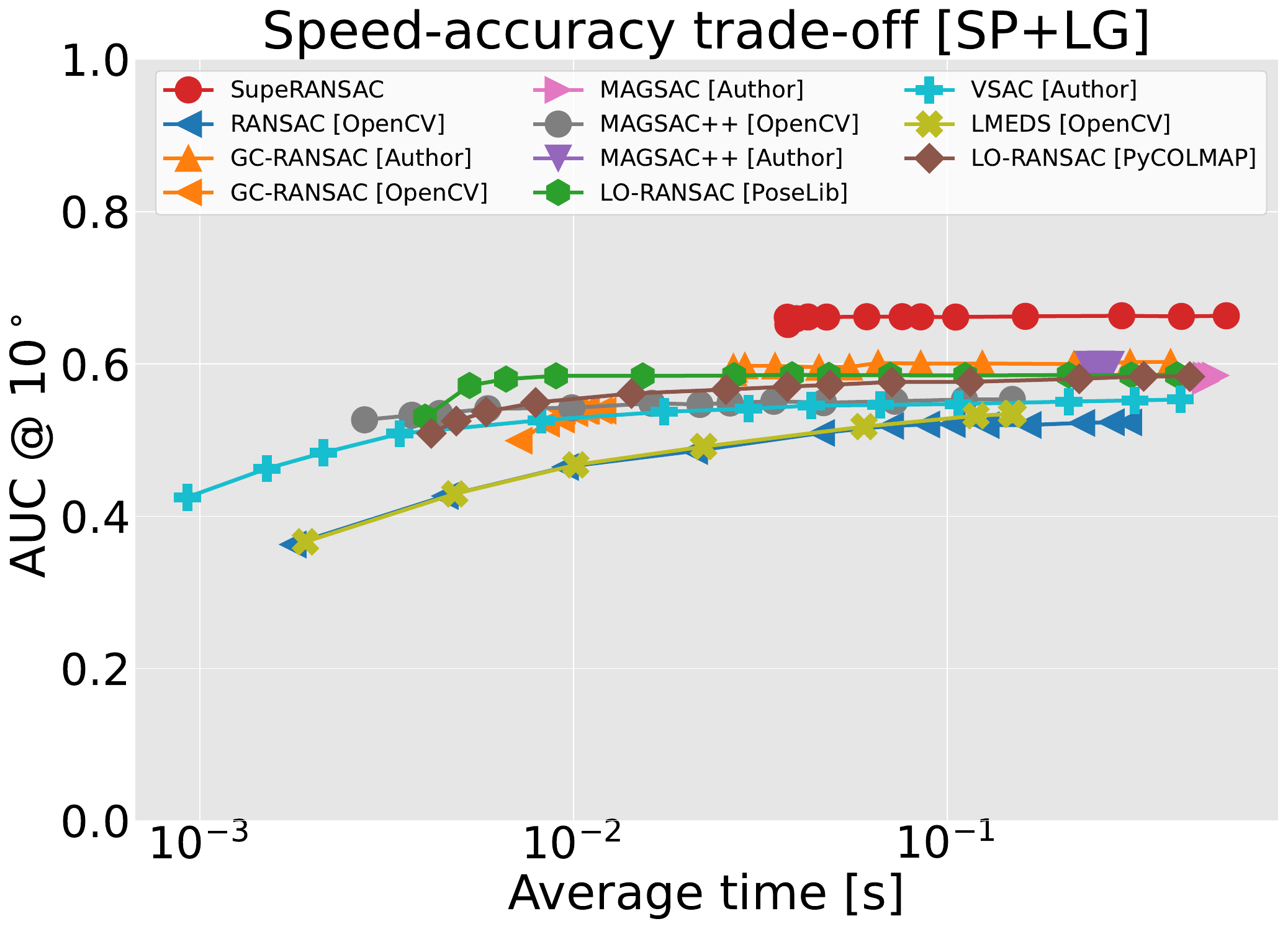}\\
        \includegraphics[width=1.0\columnwidth]{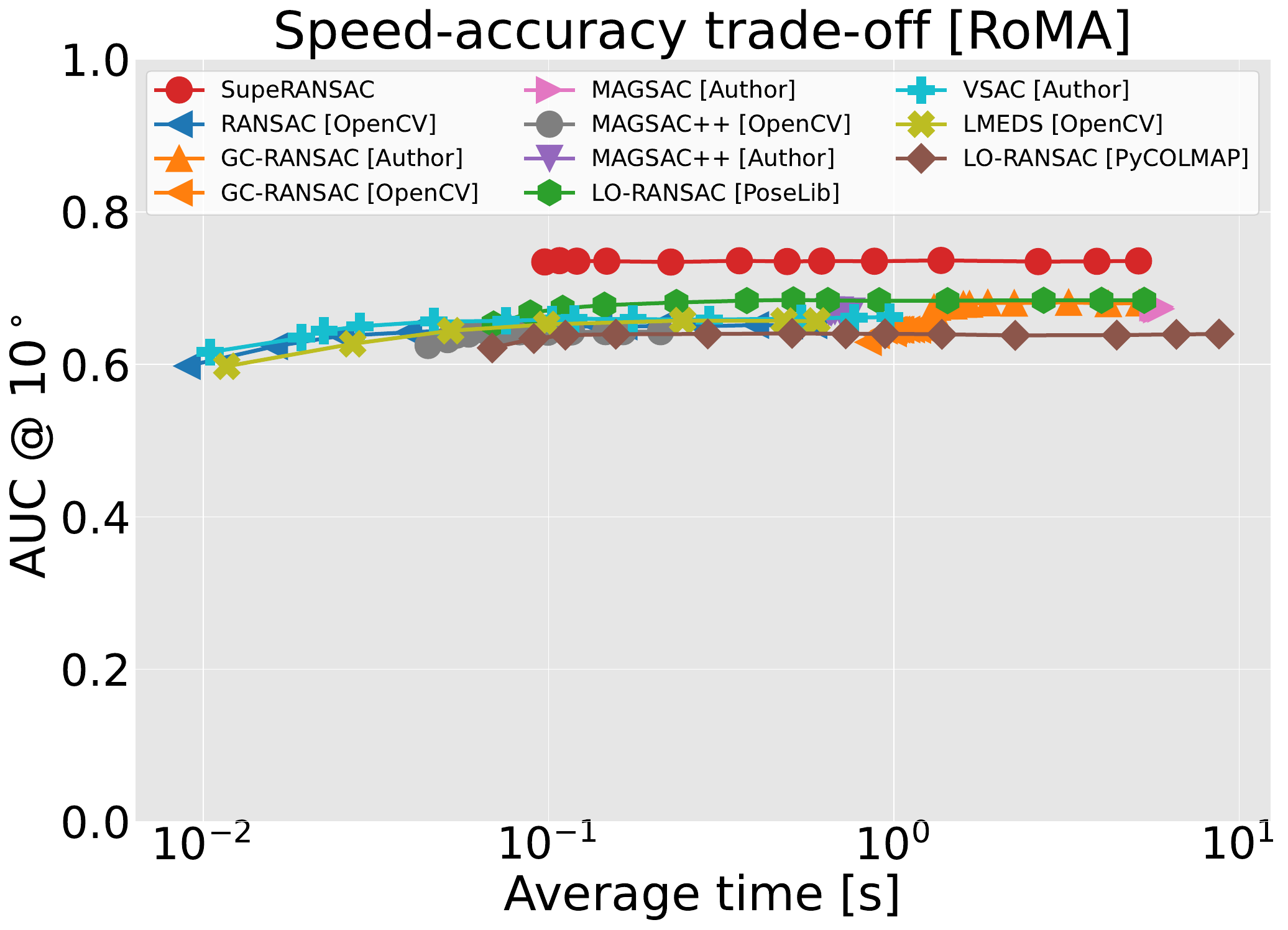}
        \caption{Essential matrix}
        \label{fig:accuracy_curves_E}
    \end{subfigure}
    ~ 
    \begin{subfigure}[t]{0.32\textwidth}
        \centering
        \includegraphics[width=1.0\columnwidth]{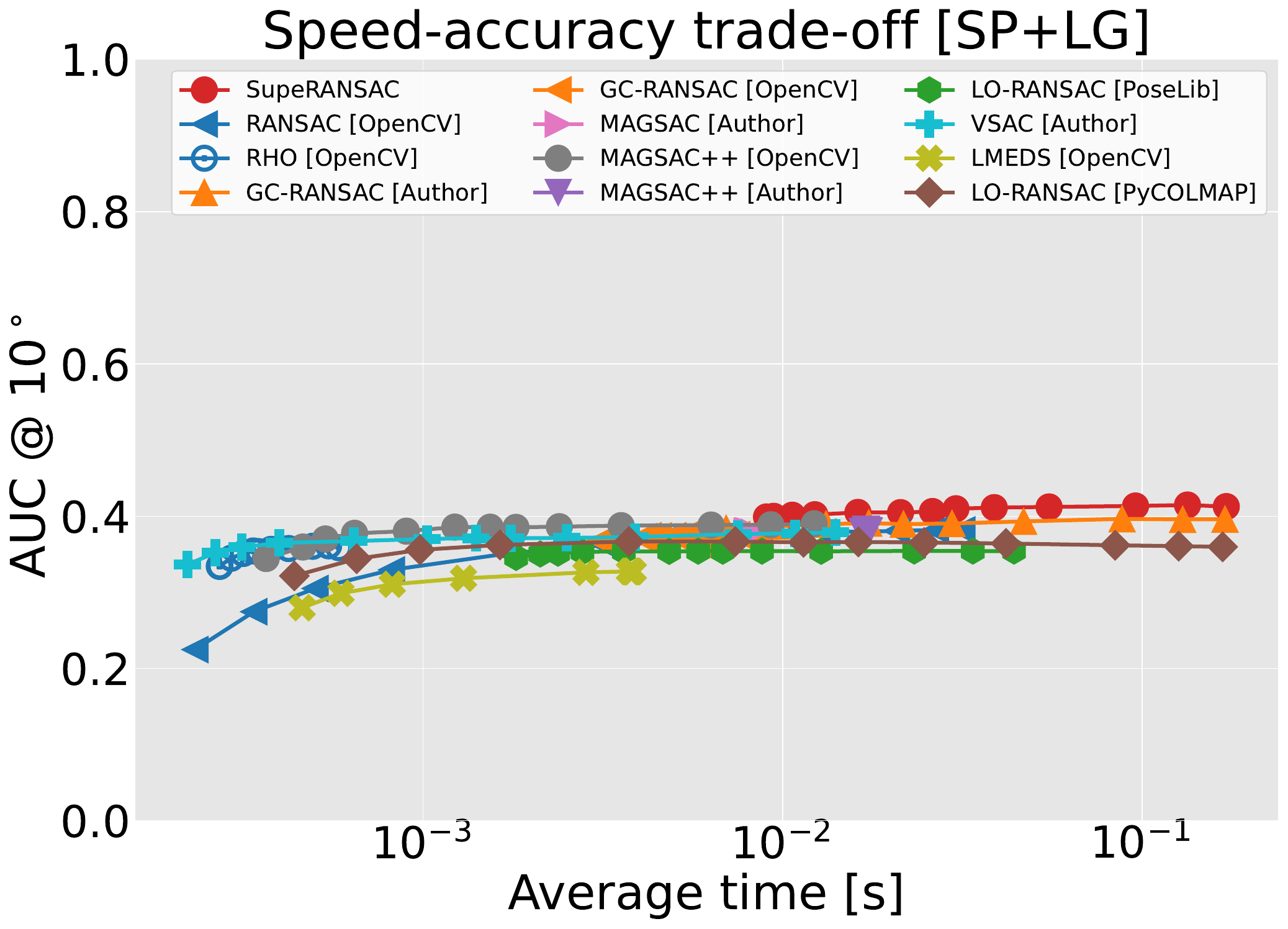}\\
        \includegraphics[width=1.0\columnwidth]{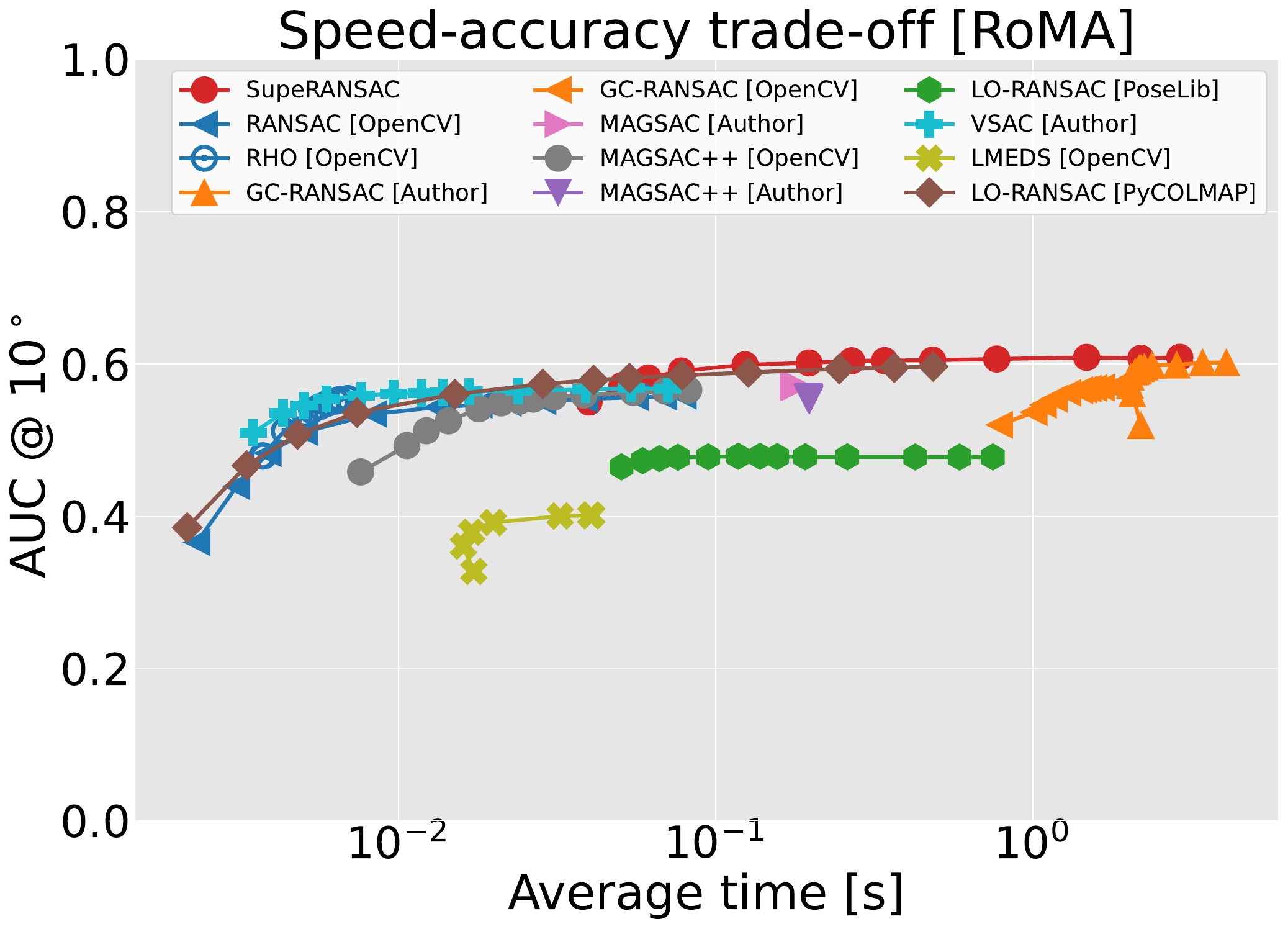}
        \caption{Homography}
        \label{fig:accuracy_curves_H}
    \end{subfigure}
    \caption{\textbf{Accuracy-runtime comparison} of robust estimators for (a) fundamental matrix, (b) essential matrix, and (c) homography estimation. 
    Each subplot displays the Area Under the recall Curve for pose error at a $10^\circ$ threshold (AUC@10$^\circ$) against the average runtime in seconds. 
    Results are averaged over six datasets: ScanNet1500~\cite{dai2017scannet,sarlin2020superglue}, PhotoTourism~\cite{snavely2006photo,jin2021image}, LaMAR~\cite{sarlin2022lamar}, 7Scenes~\cite{glocker2013real}, ETH3D~\cite{schops2019bad}, and KITTI~\cite{geiger2013vision}, totaling 39,592 image pairs. 
    The evaluation is presented for two distinct feature types: SuperPoint~\cite{detone2018superpoint}+LightGlue~\cite{lindenberger2023lightglue} correspondences (top row) and RoMA~\cite{edstedt2024roma} matches (bottom). Optimal performance is indicated by curves approaching the top-left corner, signifying high accuracy achieved with low runtime.}
    \label{fig:accuracy_curves}
\end{figure*}

\subsection{Minimal Model Estimation}

The objective of minimal model estimation is to rapidly generate model hypotheses from the smallest possible subset of data points. 
In this stage, computational efficiency is crucial, as a model is estimated for each new minimal sample selected by the RANSAC process. 
Models that pass initial validity checks (both sample and model degeneracy checks) are then typically subject to more extensive evaluation and optimization. 
The specifics of minimal model estimation for each geometric problem considered are outlined below.

\paragraph{Homography Estimation}
Given four point correspondences, $(p_1, p_1'), \dots, (p_4, p_4')$, where $p_i \leftrightarrow p_i'$, the homography matrix $H \in \mathbb{R}^{3 \times 3}$ is estimated using the Direct Linear Transform (DLT) algorithm, also known as the four-point algorithm~\cite{hartley2003multiple}. Each correspondence yields two linear inhomogeneous equations in the elements of $H$. With four correspondences, we obtain a system of eight linear equations. We employ an efficient solution by setting the bottom-right element of $H$ to 1 (\ie, $H_{33}=1$) and solving the resulting $8 \times 8$ linear system $Ah = b$ using Gauss elimination. The input correspondences are assumed to be normalized as described in our preprocessing stage (Sec.~\ref{sec:superansac}), which is crucial for the numerical stability of the DLT algorithm.

\paragraph{Fundamental Matrix Estimation}
For fundamental matrix estimation from seven point correspondences, we utilize the normalized seven-point algorithm~\cite{hartley2003multiple}. This method involves constructing a $7 \times 9$ coefficient matrix $A$ from the correspondences, where each correspondence $(x_i,y_i) \leftrightarrow (x_i',y_i')$ contributes a row $A_i \mathbf{f} = 0$ with $\mathbf{f}$ being the vectorized $F$ matrix. The solution space for $\mathbf{f}$ is the two-dimensional null space of $A$, typically found as linear combinations of two basis vectors $F_1, F_2$. We compute this null space efficiently using full-pivoting LU decomposition. The constraint $\det(F)=0$ is then imposed, leading to a cubic polynomial in terms of the linear combination parameter $\alpha$ as:
\begin{equation*}
    \det(\alpha F_1 + (1-\alpha)F_2)=0.
\end{equation*}
Solving this cubic equation yields one or three real solutions for the fundamental matrix $F$.

\paragraph{Essential Matrix Estimation}
When five point correspondences are available, the essential matrix $E \in \mathbb{R}^{3 \times 3}$ is estimated using Nistér's efficient five-point algorithm~\cite{nister2004efficient}. This algorithm first computes the four-dimensional null space of the $5 \times 9$ coefficient matrix (derived from the epipolar constraint $p_i'^T E p_i = 0$) using full-pivoting LU decomposition. It then leverages the Demazure constraints~\cite{demazure1988deux} (including $\det(E)=0$ and $2EE^TE - \text{trace}(EE^T)E = 0$) to formulate a system of polynomial equations. These are ultimately reduced to solving a tenth-degree univariate polynomial, for which we find roots using Sturm bracketing, yielding up to ten real solutions for essential matrix $E$.

\paragraph{Rigid Transformation Estimation}
Given a minimal set of three 3D-to-3D point correspondences, the rigid transformation $T \in \text{SE}(3)$ (comprising rotation $R \in \text{SO}(3)$ and translation $t \in \mathbb{R}^3$) is determined. We employ a standard Procrustes algorithm~\cite{gower2004procrustes} based on singular value decomposition to find the optimal rotation and subsequently the translation.

\paragraph{Absolute Pose Estimation}
For absolute pose estimation from three 2D-to-3D correspondences (the P3P problem), we compute the camera rotation $R \in \text{SO}(3)$ and translation $t \in \mathbb{R}^3$. SupeRANSAC utilizes the Lambda Twist P3P solver~\cite{persson2018lambda}, known for its efficiency and numerical stability.

\subsection{Model Degeneracy Checks}

Subsequent to minimal model estimation, and prior to more computationally intensive evaluations like inlier counting, model degeneracy checks are performed. These checks aim to swiftly identify and discard incorrect models that may have arisen from the minimal solver, particularly due to numerical precision issues or near-degenerate input samples that passed the initial sample degeneracy filters.

\paragraph{Homography Estimation}
A computed homography matrix $H$ is checked for near-degeneracy by examining its determinant. If $|\det(H)|$ is excessively small, it indicates that the transformation collapses areas significantly, rendering it infeasible in reality. Similarly, an extremely large determinant might also suggest numerical instability or an unreasonable transformation. We thus discard $H$ if $|\det(H)|$ falls outside a predefined plausible range $[\epsilon_{\text{low}}, \epsilon_{\text{high}}]$.

\paragraph{Rigid Transformation and Absolute Pose Estimation}
For both rigid transformation and absolute pose estimation, the estimated rotation $R$ must be a proper rotation, \ie, $R \in \text{SO}(3)$. This is verified by checking conditions: $R^T R \approx I$ (orthogonality, usually ensured by the solver) and, critically, $\det(R) \approx +1$. If $\det(R) \approx -1$, the matrix represents a reflection or an improper rotation, and the model is rejected.

\paragraph{Essential and Fundamental Matrix Estimation}
For models of epipolar geometry ($F$ and $E$ matrices), we do not apply explicit post-solution model degeneracy checks beyond those inherently enforced by the solvers (\eg, the rank-2 constraint for $F$-matrices from the 7-point algorithm, or the singular value constraints for $E$-matrices if fully enforced by the 5-point solver). 
While methods analyzing the condition or specific properties of $F$/$E$ matrices exist, and techniques like DEGENSAC~\cite{chum2005two} or QDEGSAC~\cite{frahm2006ransac} address sample degeneracies, we found that additional checks on the computed model had negligible impact on performance within our comprehensive SupeRANSAC pipeline, which includes robust sample selection and local optimization stages.

\subsection{Model Scoring}
\label{sec:model_scoring}

Once a model hypothesis is generated, its quality must be assessed by calculating a score based on how well it explains the observed data. We conducted a comparative analysis of several prominent scoring functions from the literature, including: standard inlier counting as in RANSAC~\cite{fischler1981random}, MSAC (M-estimator SAmple Consensus)~\cite{torr2000mlesac}, MAGSAC (Marginalizing SAmple Consensus)~\cite{barath2021marginalizing}, MAGSAC++~\cite{barath2020magsac++}, the scoring mechanism within VSAC (Verified SAmple Consensus)~\cite{ivashechkin2021vsac}, MINPRAN (MINimum PRobability of RANdomness)~\cite{stewart1995minpran}, and a contrario RANSAC~\cite{moisan2012automatic}.

RANSAC employs a hard inlier-outlier threshold, and its score is simply the number of inliers. 
MSAC also uses a fixed threshold but assigns a score to each inlier based on its residual (typically a truncated quadratic cost, bounded by the squared threshold), summing these scores and adding a fixed penalty for each outlier. 
This makes it less sensitive to the exact threshold value than pure inlier counting. 
In contrast, MAGSAC and its successor MAGSAC++ avoid a single hard threshold by marginalizing the likelihood of a point being an inlier over a range of noise scales (or, equivalently, thresholds), thereby providing a more robust quality measure per point. 
MINPRAN and a contrario RANSAC are statistically motivated methods that aim to estimate both the model score and an optimal data-dependent threshold by minimizing the probability that the observed consensus set could have arisen from random data. 
VSAC enhances MAGSAC++ scoring by incorporating a sophisticated verifier to identify and penalize points that might appear as inliers merely by chance.

Based on extensive empirical evaluations across a diverse set of geometric estimation problems and large-scale public datasets, we concluded that MAGSAC++~\cite{barath2020magsac++} consistently achieves the best trade-off in terms of average model accuracy and robustness to parameter choices. Consequently, SupeRANSAC adopts MAGSAC++ as its default scoring function.

\subsection{Preemptive Model Verification}
\label{sec:preemptive_verification}

To enhance computational efficiency, particularly when dealing with a large number of input data points, preemptive model verification techniques are employed. 
The goal is to terminate the score calculation for a given model hypothesis as soon as it becomes evident that it is unlikely to surpass the score of the current best model found so far. 
Notable existing strategies include the probabilistic Sequential Probability Ratio Test (SPRT)~\cite{chum2008optimal} and the more recent Space-Partitioning (SP) RANSAC~\cite{barath2022space}, the latter offering formal guarantees on not degrading solution accuracy.

In the development of SupeRANSAC, we evaluated these approaches. 
We observed that while SPRT can yield significant speed ups, this sometimes comes at the cost of reduced accuracy, especially in scenarios with low inlier ratios where premature termination can be detrimental. 
SP-RANSAC, while guaranteed not to compromise accuracy, provided notable speed-ups primarily for homography estimation in our experiments. 
Thus, SP-RANSAC is selectively employed within SupeRANSAC for homography estimation only.

For other problem types, and as a general fall-back, SupeRANSAC incorporates a simpler, yet effective, preemptive rule based on an iteratively updated upper bound on the achievable model score. Let $\mathcal{C} = \{c_1, \dots, c_n\}$ be the set of $n$ correspondences, $M$ be the current model hypothesis, and $q(M, c_j)$ be the quality contribution of correspondence $c_j$ to the model $M$ under the chosen scoring function (\eg, MAGSAC++). The score $s^{(k)}$ after evaluating the first $k$ correspondences is $s^{(k)} = \sum_{j=1}^{k} q(M, c_j)$. The maximum possible quality contribution from any single correspondence is denoted $q_{\text{max}}$ (for MAGSAC++, this is typically normalized to 1). After evaluating $k$ correspondences, the optimistic maximum score this model could achieve is:
\begin{equation*}
    s^{(k)}_{\text{optimistic}} = s^{(k)} + (n-k) \cdot q_{\text{max}}.
\end{equation*}
If $s_{\text{best}}$ is the score of the best model found so far, the evaluation of model $M$ can be safely terminated if $s^{(k)}_{\text{optimistic}} \leq s_{\text{best}}$. This strategy is broadly applicable to any scoring function where such an upper bound per point can be defined and is consistently applied in SupeRANSAC.

\begin{table}[t]
\centering
\scriptsize
\resizebox{1.0\columnwidth}{!}{
\begin{tabular}{c@{\hskip 8pt}l|ccccc@{\hskip 8pt}}
\toprule
    & Method & AUC@5$^\circ$ $\uparrow$ & @10$^\circ$ $\uparrow$ & @20$^\circ$ $\uparrow$ & med.\ $\epsilon$ $\downarrow$ & time (s) $\downarrow$ \\
\midrule
  \multirow{11}{*}{\rotatebox[origin=c]{90}{SP+LG matches~\cite{detone2018superpoint,lindenberger2023lightglue}}} & \textbf{SupeRANSAC} & \textbf{0.45} & \textbf{0.59} & \textbf{0.70} & \textbf{1.85} & 0.06 \\
    & RANSAC [OpenCV] & 0.23 & 0.38 & 0.53 & 3.93 & \textbf{0.01} \\
    & LMEDS [OpenCV] & 0.22 & 0.36 & 0.52 & 3.98 & \textbf{0.01} \\
    & GC-RSC [Author] & 0.37 & 0.51 & 0.64 & 2.36 & 0.05 \\
    & GC-RSC [OpenCV] & 0.32 & 0.46 & 0.59 & 2.75 & 0.01 \\
    & MAGSAC [Author] & 0.30 & 0.42 & 0.55 & 3.05 & 0.22 \\
    & MAGSAC++ [Author] & 0.28 & 0.41 & 0.54 & 2.94 & 0.55 \\
    & MAGSAC++ [OpenCV] & 0.33 & 0.48 & 0.61 & 2.66 & 0.01 \\
    & LO-RSC [Poselib] & 0.39 & 0.53 & 0.66 & 2.22 & 0.02 \\
    & LO-RSC [COLMAP] & 0.32 & 0.47 & 0.62 & 2.68 & 0.02 \\
    & VSAC [Author] & 0.34 & 0.49 & 0.63 & 2.49 & \textbf{0.01} \\
\midrule
  \multirow{11}{*}{\rotatebox[origin=c]{90}{RoMA matches~\cite{edstedt2024roma}}} & \textbf{SupeRANSAC} & \textbf{0.60} & \textbf{0.72} & \textbf{0.81} & \textbf{1.09} & 0.81 \\
    & RANSAC [OpenCV] & 0.41 & 0.57 & 0.70 & 1.73 & \textbf{0.02} \\
    & LMEDS [OpenCV] & 0.47 & 0.60 & 0.72 & 1.60 & 0.10 \\
    & GC-RSC [Author] & 0.51 & 0.65 & 0.77 & 1.27 & 1.80 \\
    & GC-RSC [OpenCV] & 0.46 & 0.61 & 0.72 & 1.51 & 0.49 \\
    & MAGSAC [Author] & 0.52 & 0.65 & 0.77 & 1.30 & 2.18 \\
    & MAGSAC++ [Author] & 0.52 & 0.65 & 0.77 & 1.29 & 2.38 \\
    & MAGSAC++ [OpenCV] & 0.49 & 0.63 & 0.75 & 1.36 & 0.03 \\
    & LO-RSC [Poselib] & 0.52 & 0.66 & 0.77 & 1.25 & 0.41 \\
    & LO-RSC [COLMAP] & 0.49 & 0.63 & 0.75 & 1.33 & 0.51 \\
    & VSAC [Author] & 0.51 & 0.65 & 0.77 & 1.28 & 0.08 \\
\bottomrule
\end{tabular}}
\caption{Performance evaluation for \textbf{fundamental matrix} estimation across six datasets: ScanNet1500~\cite{dai2017scannet,sarlin2020superglue}, PhotoTourism~\cite{snavely2006photo,jin2021image}, LaMAR~\cite{sarlin2022lamar}, 7Scenes~\cite{glocker2013real}, ETH3D~\cite{schops2019bad}, and KITTI~\cite{geiger2013vision}, totaling 39,592 image pairs. Results are presented for two distinct feature sets, SuperPoint+LightGlue (SP+LG)~\cite{detone2018superpoint,lindenberger2023lightglue} and RoMA~\cite{edstedt2024roma} matches. 
All compared methods ran for a fixed 1000 iterations. 
We report the Area Under the Curve (AUC) for relative pose error at $5^\circ, 10^\circ,$ and $20^\circ$ thresholds, the median relative pose error ($\text{med }\epsilon$) in degrees, and the average runtime per image pair in seconds.}
\label{tab:fundamental_matrix_estimation}
\end{table}

\subsection{Model Optimization}
Model optimization, often referred to as local optimization (LO), is a pivotal step for achieving high-accuracy geometric estimates~\cite{chum2003locally,lebeda2012fixing}. This process refines an initial model hypothesis, typically derived from a minimal sample, by leveraging the set of correspondences identified as its inliers.

LO-RANSAC~\cite{chum2003locally} introduced several effective strategies for this refinement, including straightforward least-squares (LS) fitting to the inlier set, iteratively reweighted least-squares (IRLS) fitting, and a ``nested RANSAC'' approach. The nested RANSAC strategy involves drawing random samples, larger than minimal, exclusively from the current inlier pool. If a model derived from such a sample yields a higher score, the inlier set is updated, and the process iterates, typically for a fixed number of iterations (\eg, 20--50).

A prominent advancement in local optimization is GC-RANSAC~\cite{barath2018graph}, which enhances the nested RANSAC paradigm by incorporating spatial coherence. This method is predicated on the observation that in many real-world scenarios, inlier correspondences tend to form spatially coherent structures. 
Thus, if a point is an inlier, its spatial neighbors are also likely to be inliers. GC-RANSAC employs a graph-cut algorithm to partition points into inliers and outliers, considering both their residuals to the current model and their neighborhood relationships. Subsequently, a nested RANSAC procedure, drawing larger-than-minimal samples, is applied to the inlier set selected by graph-cut to further refine the model. For efficient neighborhood determination within GC-RANSAC, particularly for 2D image correspondences, we utilize a 4D uniform grid structure (\eg, based on source and destination image coordinates).

\begin{table}[t]
\centering
\scriptsize
\resizebox{1.0\columnwidth}{!}{
\begin{tabular}{c@{\hskip 8pt}l|ccccc@{\hskip 8pt}}
\toprule
    & Method & AUC@5$^\circ$ $\uparrow$ & @10$^\circ$ $\uparrow$ & @20$^\circ$ $\uparrow$ & med $\epsilon$ $\downarrow$ & time (s) $\downarrow$ \\
\midrule
  \multirow{11}{*}{\rotatebox[origin=c]{90}{SP+LG matches~\cite{detone2018superpoint,lindenberger2023lightglue}}} & \textbf{SupeRANSAC} & \textbf{0.53} & \textbf{0.66} & \textbf{0.76} & \textbf{1.37} & 0.08 \\
    & RANSAC [OpenCV] & 0.36 & 0.52 & 0.66 & 2.24 & 0.10 \\
    & LMEDS [OpenCV] & 0.38 & 0.53 & 0.67 & 2.13 & 0.15 \\
    & GC-RSC [Author] & {0.46} & {0.60} & {0.73} & 1.72 & 0.07 \\
    & GC-RSC [OpenCV] & 0.41 & 0.54 & 0.67 & 2.09 & \textbf{0.01} \\
    & MAGSAC [Author] & 0.43 & 0.58 & 0.72 & 1.78 & 0.52 \\
    & MAGSAC++ [Author] & 0.45 & 0.60 & 0.73 & 1.67 & 0.26 \\
    & MAGSAC++ [OpenCV] & 0.41 & 0.55 & 0.68 & 2.00 & 0.03 \\
    & LO-RSC [Poselib] & 0.45 & 0.59 & 0.70 & 1.70 & 0.05 \\
    & LO-RSC [COLMAP] & 0.42 & 0.57 & 0.71 & 1.88 & 0.05 \\
    & VSAC [Author] & 0.40 & 0.55 & 0.69 & 2.16 & 0.04 \\
\midrule
  \multirow{11}{*}{\rotatebox[origin=c]{90}{RoMA matches~\cite{edstedt2024roma}}} & \textbf{SupeRANSAC} & \textbf{0.61} & \textbf{0.74} & \textbf{0.83} & \textbf{1.04} & 0.62 \\
    & RANSAC [OpenCV] & 0.51 & 0.65 & 0.76 & 1.32 & 0.17 \\
    & LMEDS [OpenCV] & 0.52 & 0.66 & 0.77 & 1.29 & 0.60 \\
    & GC-RSC [Author] & 0.54 & 0.68 & 0.79 & 1.20 & 1.66 \\
    & GC-RSC [OpenCV] & 0.51 & 0.65 & 0.76 & 1.26 & 1.04 \\
    & MAGSAC [Author] & 0.53 & 0.67 & 0.79 & 1.25 & 5.93 \\
    & MAGSAC++ [Author] & 0.53 & 0.67 & 0.78 & 1.26 & 0.66 \\
    & MAGSAC++ [OpenCV] & 0.50 & 0.64 & 0.76 & 1.30 & \textbf{0.09} \\
    & LO-RSC [Poselib] & 0.55 & 0.68 & 0.80 & 1.19 & 0.64 \\
    & LO-RSC [COLMAP] & 0.49 & 0.64 & 0.77 & 1.29 & 0.94 \\
    & VSAC [Author] & 0.52 & 0.66 & 0.78 & 1.24 & 0.12 \\
\bottomrule
\end{tabular}}
\caption{Performance evaluation for \textbf{essential matrix} estimation across six datasets: ScanNet1500~\cite{dai2017scannet,sarlin2020superglue}, PhotoTourism~\cite{snavely2006photo,jin2021image}, LaMAR~\cite{sarlin2022lamar}, 7Scenes~\cite{glocker2013real}, ETH3D~\cite{schops2019bad}, and KITTI~\cite{geiger2013vision}, totaling 39,592 image pairs. Results are presented for two distinct feature sets, SuperPoint+LightGlue (SP+LG)~\cite{detone2018superpoint,lindenberger2023lightglue} and RoMA~\cite{edstedt2024roma} matches. 
All compared methods ran for a fixed 1000 iterations. 
We report the Area Under the Curve (AUC) for relative pose error at $5^\circ, 10^\circ,$ and $20^\circ$ thresholds, the median relative pose error ($\text{med }\epsilon$) in degrees, and the average runtime per image pair in seconds.}
\label{tab:essential_matrix_estimation}
\end{table}

Our SupeRANSAC pipeline distinguishes between two types of model optimization: local optimization (LO), performed iteratively during the main RANSAC loop, and final optimization (FO), executed once on the best model found. This distinction is motivated by their differing objectives: LO prioritizes \textit{efficient} model improvement to guide the search, whereas FO can afford more computationally intensive refinement as it is applied only to the final candidate model.

\paragraph{Local Optimization (LO)}
For local optimization, SupeRANSAC defaults to GC-RANSAC when the number of correspondences is manageable (\eg, up to a few thousand). However, the computational cost of graph-cut can become prohibitive for very large datasets, such as those arising from dense image matching or large point clouds. Therefore, if the number of correspondences exceeds a threshold (empirically set to 2000 in our implementation), SupeRANSAC automatically transitions to the nested RANSAC strategy for LO. In each iteration of this nested RANSAC, we draw samples of size $7m$, where $m$ is the minimal sample size required for the specific geometric model, following~\cite{lebeda2012fixing}.

\paragraph{Final Optimization (FO)}
For the final optimization stage, SupeRANSAC employs an iteratively reweighted least-squares (IRLS) approach. This involves using robust Cauchy weights~\cite{fattorini1983cauchy} applied to the residuals, with a strategy of iteratively halving the inlier threshold (for determining the active set in IRLS) and re-evaluating inlier consensus at each iteration of the FO process.

\subsection{Nonminimal Model Estimation}
\label{sec:nonminimal_estimation}

Nonminimal model estimation refers to the process of computing model parameters from a set of correspondences larger than the minimal required size. This is a core component of both the local optimization (LO) and final optimization (FO) stages. While accuracy is paramount, computational efficiency remains a consideration, though less critical than for minimal solvers as these routines are executed less frequently. The specific nonminimal solvers employed for each estimation problem are detailed below.

\paragraph{Homography Estimation}
For estimating a homography from a nonminimal set of correspondences, we utilize the standard normalized Direct Linear Transformation (DLT) algorithm~\cite{hartley2003multiple}. Input points are first normalized to improve numerical stability~\cite{hartley1997defense}. The DLT algorithm then provides a closed-form solution by solving a linear system of equations derived from all inlier correspondences. In our pipeline, this typically provides sufficient accuracy without requiring further iterative non-linear refinement for this specific model.

\begin{figure}[t!]
    \centering
    \includegraphics[width=0.9\columnwidth]{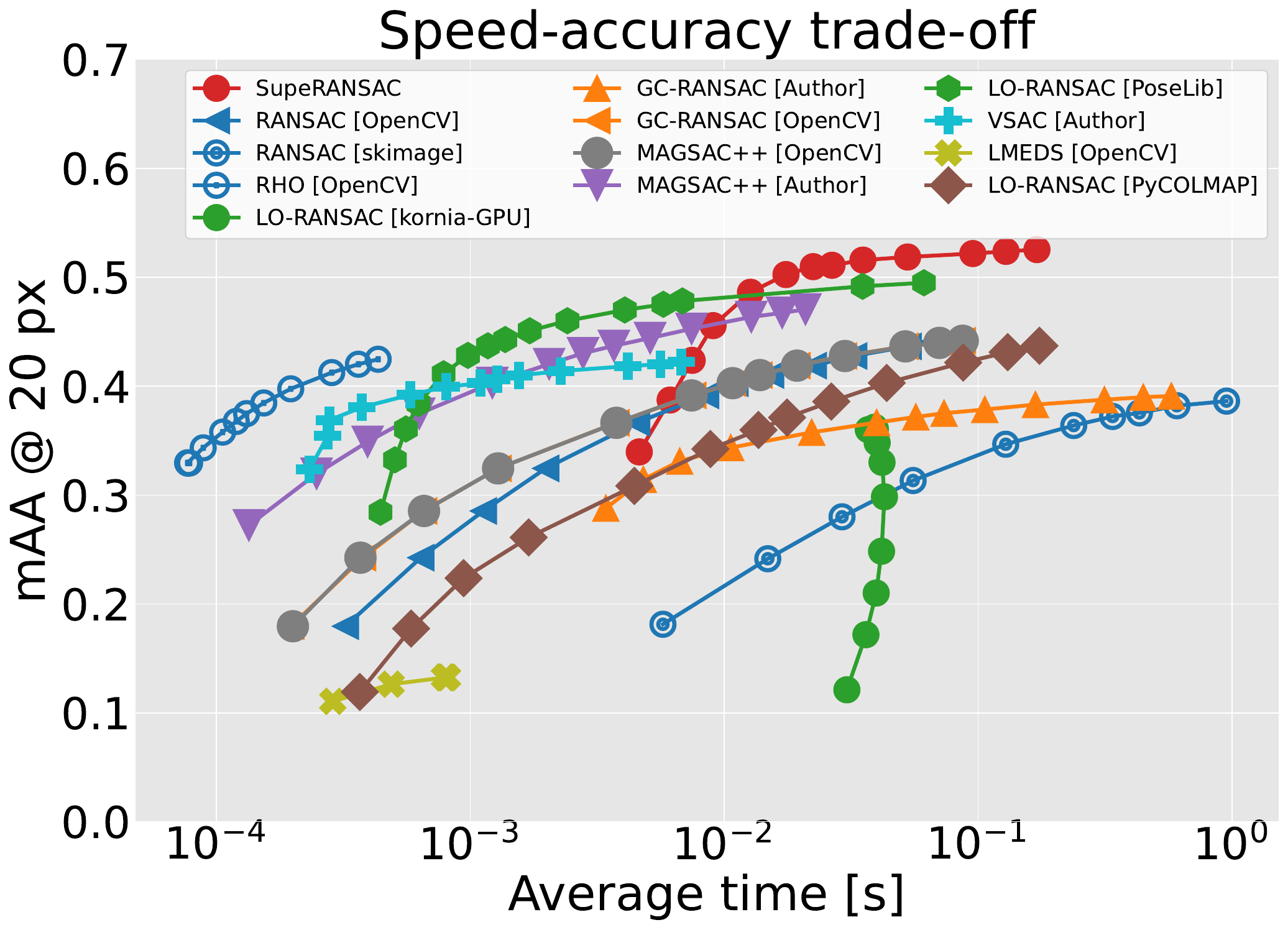}
    \caption{Mean Average Accuracy (mAA) scores versus runtime (secs) for \textbf{homography} estimation on the HEB dataset~\cite{barath2023large}. Curves approaching the top-left corner indicate superior performance, achieving both high accuracy and low runtime.}
    \label{fig:homography_heb_curve}
\end{figure}

\paragraph{Essential Matrix Estimation}
To estimate an essential matrix from $n > 5$ correspondences, we first obtain an initial algebraic estimate. This is achieved by forming an $n \times 9$ coefficient matrix from the epipolar constraint $p_i'^T E p_i = 0$ for all $n$ correspondences (assuming normalized image coordinates). The vectorized essential matrix $\mathbf{e}$ is then found as the right singular vector corresponding to the smallest singular value of this coefficient matrix (analogous to the normalized eight-point algorithm~\cite{hartley1997defense} but applied for $E$). We found this approach to be more consistently accurate for generating the initial nonminimal estimate than directly using a minimal 5-point solver (such as~\cite{nister2004efficient} or~\cite{stewenius2008minimal}) on $N>5$ points or its direct algebraic extensions without proper handling of the non-minimal case. From the resulting $E$ matrix, up to four potential rotation and translation solutions are decomposed, and the physically correct one is selected using the chirality constraint (ensuring points are in front of both cameras)~\cite{hartley2003multiple}.

Following this algebraic initialization, we perform a non-linear refinement using the Levenberg-Marquardt (LM) algorithm. This optimization directly refines the five degrees of freedom of the relative pose (parameterized, for instance, by a rotation $R \in \text{SO}(3)$ and a unit translation vector $t \in \mathbb{R}^3, \|t\|=1$) by minimizing the sum of squared Sampson errors. During this LM optimization, residuals are weighted using the MAGSAC++ weighting scheme~\cite{barath2020magsac++}. If multiple essential matrix hypotheses were propagated (\eg, from different RANSAC iterations leading to LO), each is refined, and the one yielding the best score post-refinement is selected.

\paragraph{Fundamental Matrix Estimation}
For nonminimal fundamental matrix estimation, an initial estimate is obtained using the normalized eight-point algorithm~\cite{hartley1997defense} (or the normalized seven-point algorithm if exactly seven correspondences are available). Subsequently, this initial estimate is refined using the Levenberg-Marquardt algorithm. To ensure the rank-2 constraint is maintained throughout the optimization, we employ the parameterization proposed by Bartoli and Sturm~\cite{bartoli2004nonlinear}:
\begin{equation}
    F = U \, \text{diag}(1, \sigma, 0) \, V^\top,
\end{equation}
where $U, V \in \text{SO}(3)$ are rotation matrices (parameterized, \eg, by unit quaternions $q_U, q_V$) and $\sigma \ge 0$ is a scalar. The optimization minimizes a robust MAGSAC++-weighted sum of squared Sampson errors by adjusting the parameters $q_U, q_V$, and $\sigma$. The final fundamental matrix is selected from any refined candidates based on its MAGSAC++ score.

\paragraph{Absolute Pose Estimation}
Initial absolute pose estimates from $n>3$ 2D-to-3D correspondences are obtained using the Efficient P$n$P (EP$n$P) solver~\cite{lepetit2009ep}. Similar to the relative pose case, these initial estimates are then refined via Levenberg-Marquardt optimization. The LM algorithm minimizes the sum of squared reprojection errors, with residuals weighted according to the MAGSAC++ scheme. This iterative refinement adjusts the six degrees of freedom of the camera pose (rotation $R \in \text{SO}(3)$ and translation $t \in \mathbb{R}^3$).

\paragraph{Rigid Transformation Estimation}
For estimating a rigid transformation from $n>3$ 3D-to-3D correspondences, the standard Procrustes algorithm (\eg, based on SVD as in~\cite{gower2004procrustes}) is employed. This method provides a closed-form, optimal least-squares solution for the rotation and translation.

\begin{table}[t]
\centering
\scriptsize
\resizebox{1.0\columnwidth}{!}{
\begin{tabular}{c@{\hskip 8pt}l|ccccc@{\hskip 8pt}}
\toprule
    & Method & AUC@5$^\circ$ $\uparrow$ & @10$^\circ$ $\uparrow$ & @20$^\circ$ $\uparrow$ & med $\epsilon$ $\downarrow$ & time (s) $\downarrow$ \\
\midrule
  \multirow{12}{*}{\rotatebox[origin=c]{90}{SP+LG matches~\cite{detone2018superpoint,lindenberger2023lightglue}}} & \textbf{SupeRANSAC} & \textbf{0.26} & \textbf{0.41} & \textbf{0.58} & \textbf{3.36} & 0.030 \\
    & RANSAC [OpenCV] & 0.23 & 0.37 & 0.54 & 3.94 & 0.006 \\
    & LMEDS [OpenCV] & 0.17 & 0.33 & 0.51 & 4.58 & 0.004 \\
    & RHO [OpenCV] & 0.20 & 0.35 & 0.53 & 4.00 & \textbf{0.000} \\
    & GC-RSC [Author] & 0.24 & 0.39 & 0.56 & 3.51 & 0.022 \\
    & GC-RSC [OpenCV] & 0.23 & 0.38 & 0.55 & 3.84 & 0.006 \\
    & MAGSAC [Author] & 0.23 & 0.38 & 0.55 & 3.59 & 0.008 \\
    & MAGSAC++ [Author] & 0.23 & 0.38 & 0.55 & 3.71 & 0.017 \\
    & MAGSAC++ [OpenCV] & 0.23 & 0.39 & 0.56 & 3.62 & 0.002 \\
    & LO-RSC [Poselib] & 0.19 & 0.35 & 0.54 & 3.84 & 0.007 \\
    & LO-RSC [COLMAP] & 0.21 & 0.37 & 0.54 & 3.88 & 0.016 \\
    & VSAC [Author] & 0.23 & 0.37 & 0.53 & 3.75 & 0.002 \\
\midrule
  \multirow{12}{*}{\rotatebox[origin=c]{90}{RoMA matches~\cite{edstedt2024roma}}} & \textbf{SupeRANSAC} & \textbf{0.46} & \textbf{0.60} & \textbf{0.74} & \textbf{1.71} & 0.341 \\
    & RANSAC [OpenCV] & 0.39 & 0.55 & 0.70 & 2.04 & 0.022 \\
    & LMEDS [OpenCV] & 0.23 & 0.40 & 0.58 & 3.57 & 0.041 \\
    & RHO [OpenCV] & 0.38 & 0.54 & 0.69 & 2.15 & \textbf{0.005} \\
    & GC-RSC [Author] & 0.45 & 0.59 & 0.73 & 1.84 & 2.192 \\
    & GC-RSC [OpenCV] & 0.43 & 0.57 & 0.70 & 2.10 & 1.492 \\
    & MAGSAC [Author] & 0.43 & 0.57 & 0.71 & 1.99 & 0.177 \\
    & MAGSAC++ [Author] & 0.40 & 0.55 & 0.70 & 2.06 & 0.197 \\
    & MAGSAC++ [OpenCV] & 0.39 & 0.55 & 0.71 & 2.00 & 0.027 \\
    & LO-RSC [Poselib] & 0.30 & 0.48 & 0.65 & 2.44 & 0.156 \\
    & LO-RSC [COLMAP] & 0.44 & 0.58 & 0.71 & 2.03 & 0.053 \\
    & VSAC [Author] & 0.42 & 0.56 & 0.70 & 2.18 & 0.014 \\
\bottomrule
\end{tabular}}
\caption{Performance evaluation for \textbf{homography} estimation across six datasets: ScanNet1500~\cite{dai2017scannet,sarlin2020superglue}, PhotoTourism~\cite{snavely2006photo,jin2021image}, LaMAR~\cite{sarlin2022lamar}, 7Scenes~\cite{glocker2013real}, ETH3D~\cite{schops2019bad}, and KITTI~\cite{geiger2013vision}, totaling 39,592 image pairs. Results are presented for two distinct feature sets, SuperPoint+LightGlue (SP+LG)~\cite{detone2018superpoint,lindenberger2023lightglue} and RoMA~\cite{edstedt2024roma} matches. 
All compared methods ran for a fixed 1000 iterations. 
We report the Area Under the Curve (AUC) for relative pose error at $5^\circ, 10^\circ,$ and $20^\circ$ thresholds, the median relative pose error ($\text{med }\epsilon$) in degrees, and the average runtime per image pair in seconds.}
\label{tab:homography_estimation}
\end{table}

\begin{table}[t]
\centering
\scriptsize
\begin{tabular}{@{\hskip 8pt}l|cccc@{\hskip 8pt}}
\toprule
    Method & mAA $\uparrow$ & med $\epsilon$ $\downarrow$ & \# inliers $\uparrow$ & time (s) $\downarrow$ \\
\midrule
    \textbf{SupeRANSAC} & \textbf{0.51} & \textbf{0.07} & \phantom{1}81 & 0.027 \\
    RANSAC [OpenCV] & 0.41 & 0.15 & \phantom{1}52 & 0.015 \\
    LMEDS [OpenCV] & 0.13 & 1.79 & \textbf{127} & 0.001 \\
    RHO [OpenCV] & 0.37 & 0.22 & \phantom{1}50 & \textbf{0.000} \\
    GC-RSC [Author] & 0.38 & 0.12 & \phantom{1}66 & 0.073 \\
    GC-RSC [OpenCV] & 0.41 & 0.15 & \phantom{1}52 & 0.014 \\
    MAGSAC++ [Author] & 0.44 & 0.13 & \phantom{1}95 & 0.004 \\
    MAGSAC++ [OpenCV] & 0.41 & 0.15 & \phantom{1}52 & 0.014 \\
    LO-RSC [Poselib] & 0.44 & 0.16 & \phantom{1}49 & 0.001 \\
    LO-RSC [COLMAP] & 0.37 & 0.33 & \phantom{1}48 & 0.018 \\
    LO-RSC [kornia] & 0.36 & 0.33 & \phantom{1}44 & 0.037 \\
    VSAC [Author] & 0.41 & 0.12 & \phantom{1}92 & 0.001 \\
\bottomrule
\end{tabular}
\caption{Performance evaluation of \textbf{homography} estimation on the HEB dataset~\cite{barath2023large}, utilizing RootSIFT correspondences~\cite{lowe2004sift} established by mutual nearest neighbor (MNN) matching. All compared robust estimators ran for a fixed 1000 iterations. The reported metrics include: mean Average Accuracy (mAA), median geometric error ($\text{med }\epsilon$) in pixels, the average number of estimated inliers (\# inliers), and the average runtime per image pair in seconds.}
\label{tab:homography_heb}
\end{table}

\subsection{Termination Criterion}
\label{sec:termination_criterion}

SupeRANSAC employs the standard RANSAC termination criterion~\cite{fischler1981random}. The iterative process stops when the probability of finding a model with a larger consensus set than the current best falls below a user-defined threshold (\eg, 1\%), given a specified confidence (\eg, 99.9\%) in having found a valid solution if one exists according to the current best inlier ratio.

\subsection{Summary}
\label{sec:summary}

SupeRANSAC is a comprehensive and highly optimized framework for robust geometric model estimation. 
It systematically integrates a suite of advanced techniques, including guided sampling strategies (PROSAC and P-NAPSAC), meticulous sample and model degeneracy checks, efficient preemptive model verification, and a sophisticated multi-stage optimization process. 
Key algorithmic choices, such as MAGSAC++ for robust model scoring and GC-RANSAC for effective local optimization, are complemented by carefully designed problem-specific nonminimal solvers and refinement procedures, collectively contributing to the state-of-the-art performance of the framework in terms of accuracy and efficiency.

\section{Experiments}

In this section, we present a thorough empirical evaluation of SupeRANSAC across a diverse range of geometric vision problems, utilizing several large-scale public datasets. We aim to demonstrate its performance in terms of accuracy, speed, and robustness compared with baselines.

\paragraph{Evaluation Metrics}

For epipolar geometry estimation (fundamental and essential matrices), we adhere to the Image Matching Challenge (IMC) benchmark protocols~\cite{jin2021image}. The primary metric is the angular relative pose error, defined as the maximum of the errors in rotation (angular difference between estimated and ground truth rotation matrices) and translation (angle between estimated and ground truth translation vectors).
For homography estimation, in addition to the epipolar geometry metrics (when decomposing H into pose), we also report the absolute pose error derived from the homography, as suggested by~\cite{barath2023large}, which provides a task-relevant measure of accuracy.
Pose errors for absolute pose estimation (P$n$P) and rigid transformation estimation (3D point cloud registration) are computed following established methodologies from prior benchmark works~\cite{sattler2018benchmarking,zeng20173dmatch}, typically involving errors in rotation and translation.
To summarize performance over varying error thresholds, we report the Area Under the recall Curve (AUC) up to 10$^\circ$ pose error~\cite{jin2021image}. For homography evaluation, particularly on the HEB dataset, we also report mean Average Accuracy (mAA) scores~\cite{barath2023large}.

\begin{table}[t]
\centering
\scriptsize
\resizebox{1.0\columnwidth}{!}{
\begin{tabular}{c@{\hskip 8pt}l|cc@{\hskip 8pt}}
\toprule
     & Method & Day & Night \\
\midrule
  \multirow{8}{*}{\rotatebox[origin=c]{90}{Aachen Day-Night}} & \textbf{SupeRANSAC} & 80.7 / 93.6 / \textbf{97.8} & \textbf{78.5} / \textbf{89.5} / \textbf{97.4} \\
   & LO-RANSAC [COLMAP] & \textbf{88.5} / \textbf{94.9} / \textbf{97.8} & 	70.2 / 88.0 / \textbf{97.4} \\
  & LO-RANSAC [PoseLib] & 88.1 / 94.3 / \textbf{97.8} & 70.7 / 87.4 / \textbf{97.4} \\
  & GC-RANSAC [Author] & 	76.7 / 89.4 / 97.5 & 59.7 / 82.2 / 96.9 \\
  & RANSAC [OpenCV-LM] & 69.2 / 88.7 / 96.6 & 71.7 / 86.4 / 95.3 \\
  & RANSAC [OpenCV-EPnP] & 60.6 / 84.7 / 96.6 & 67.0 / 87.4 / 95.3 \\
  & RANSAC [OpenCV-P3P] & 61.0 / 83.5 / 97.0 & 65.4 / 83.8 / 96.9 \\
  & RANSAC [OpenCV-SQPnP] & 67.2 / 88.5 / 96.6 & 71.2 / 86.9 / 95.3 \\
\midrule
     & Method & DUC1 & DUC2 \\
\midrule
  \multirow{8}{*}{\rotatebox[origin=c]{90}{InLoc}} & \textbf{SupeRANSAC} & \textbf{46.0} / \textbf{67.2} / \textbf{77.3} & \textbf{53.4} / \textbf{70.2} / \textbf{73.3} \\
   & LO-RANSAC [COLMAP] & 43.9 / 65.2 / 75.3 & 44.3 / 67.9 / \textbf{73.3} \\
  & LO-RANSAC [PoseLib] & 44.4 / 66.2 / 76.3 & 50.4 / \textbf{70.2} / \textbf{73.3} \\
  & GC-RANSAC [Author] & 39.4 / 61.6 / 74.7 & 35.9 / 58.0 / 67.2 \\
  & RANSAC [OpenCV-LM] & 27.3 / 39.9 / 51.0 & 22.1 / 35.1 / 41.2 \\
  & RANSAC [OpenCV-EPnP] & 22.2 / 37.9 / 50.5 & 19.8 / 35.1 / 42.0 \\
  & RANSAC [OpenCV-P3P] & 31.3 / 48.0 / 63.1 & 29.8 / 45.8 / 54.2 \\
  & RANSAC [OpenCV-SQPnP] & 25.3 / 38.9 / 52.5 & 21.4 / 35.9 / 42.0 \\
\bottomrule
\end{tabular}}
\caption{Comparative results for \textbf{absolute camera pose} estimation on the Aachen Day-Night~\cite{sattler2018benchmarking} and InLoc~\cite{taira2018inloc} visual localization benchmarks. The evaluation employs SuperPoint+LightGlue (SP+LG)~\cite{detone2018superpoint,lindenberger2023lightglue} correspondences, with various robust estimators integrated into the HLoc pipeline~\cite{sarlin2019coarse}. Performance is quantified by Area Under the Curve (AUC) scores at three distinct precision thresholds for each dataset: (0.25m, 2$^\circ$), (0.5m, 5$^\circ$), and (5m, 10$^\circ$) for Aachen Day-Night; and (0.25m, 10$^\circ$), (0.5m, 10$^\circ$), and (1m, 10$^\circ$) for InLoc.}
\label{tab:absolute_pose}
\end{table}

\paragraph{Feature Correspondences}
To demonstrate broad applicability, experiments involving 2D image data (epipolar geometry, homography, absolute pose) utilize correspondences generated by two distinct state-of-the-art methods:
(1) SuperPoint features~\cite{detone2018superpoint} matched with LightGlue~\cite{lindenberger2023lightglue}, representing high-quality sparse correspondences.
(2) RoMA~\cite{edstedt2024roma}, providing dense, robust optical flow fields from which correspondences are derived.
Additionally, for specific homography estimation benchmarks (HEB dataset), we include experiments using traditional RootSIFT features~\cite{lowe2004sift} with mutual nearest neighbor (MNN) matching to assess performance with classic feature types.
For rigid transformation estimation between 3D point clouds, we use pre-computed correspondences from GeoTransformer~\cite{qin2023geotransformer}.

\paragraph{Parameter Tuning}
Parameters for all compared robust estimators were tuned fairly. For epipolar geometry estimation, we randomly selected 200 image pairs from each dataset to form a small tuning set. Key parameters, such as the inlier-outlier error threshold and, for methods like GC-RANSAC, the spatial coherence weight, were optimized on this set. These tuning pairs were not subsequently removed from the test set; given the limited number of parameters (typically 1-2) in robust estimators, overfitting to such a small subset is highly unlikely and not observed in practice.
For homography estimation, parameters were tuned using the designated training split of the HEB dataset~\cite{barath2023large}.
For rigid transformation estimation, a single, randomly selected scene from the 3DMatch dataset~\cite{zeng20173dmatch} was used for tuning.
For absolute pose estimation, we adopted the default inlier-outlier threshold provided by the HLoc framework~\cite{sarlin2019coarse}, avoiding specific tuning for this task.
Crucially, once parameters were determined for a given problem type and estimator, they remained fixed across all test datasets for that problem, ensuring a fair comparison.

\paragraph{Baselines}
SupeRANSAC is compared with a comprehensive set of state-of-the-art robust estimation algorithms.
For homography, epipolar geometry, and absolute pose estimation, baselines include implementations from OpenCV~\cite{bradski2000opencv} (RANSAC, LMedS, GC-RANSAC, MAGSAC++, and RHO for homographies), PoseLib~\cite{poselib} (which implements an LO-RANSAC variant), and pyCOLMAP~\cite{schonberger2016structure} (also featuring an LO-RANSAC approach). We further compare against standalone implementations of GC-RANSAC~\cite{barath2018graph}, VSAC~\cite{ivashechkin2021vsac}, MAGSAC, and MAGSAC++~\cite{barath2021marginalizing,barath2020magsac++}. While other libraries like scikit-image or Kornia exist, they were not included as primary baselines due to preliminary evaluations indicating significantly slower Python-based execution or limitations in accuracy for these specific tasks compared to the selected C++ based or highly optimized libraries.
For rigid transformation estimation, comparisons are made against RANSAC implementations in OpenCV and Open3D~\cite{zhou2018open3d}, as well as GC-RANSAC, MAGSAC, MAGSAC++, and a standard iteratively reweighted Procrustes algorithm.

\subsection{Fundamental Matrix Estimation}

We evaluate fundamental matrix estimation on diverse real-world datasets, encompassing indoor and outdoor scenes with varying ground truth (GT) acquisition methods.
The 7Scenes dataset~\cite{glocker2013real} provides RGB-D sequences of seven indoor environments. 
We select pairs by sampling every 10th image $i$ and pairing it with image $i+50$ from the test sequences (1600 pairs).
ScanNet~\cite{dai2017scannet} is a large-scale indoor RGB-D dataset. 
We use the standard test set of 1500 images following~\cite{sarlin2020superglue}.
The PhotoTourism dataset~\cite{snavely2006photo} consists of large-scale outdoor SfM reconstructions. 
We use the 9900 validation pairs from the CVPR IMC 2020~\cite{jin2021image}.
ETH3D~\cite{schops2019bad} offers high-resolution multi-view imagery of indoor and outdoor scenes. 
We sample pairs from the 13 training scenes that share at least 500 GT keypoints (1969 pairs).
The KITTI Visual Odometry dataset~\cite{geiger2013vision} features driving scenarios.
We form 23,190 pairs from consecutive frames in the 11 training sequences.
Finally, the LaMAR dataset~\cite{sarlin2022lamar}, designed for augmented reality research, provides indoor/outdoor sequences.
We use 1423 consecutive image pairs from the HoloLens validation split in the CAB building.
In total, our fundamental matrix evaluation spans 39,592 image pairs.
The reported results are averaged across the datasets, ensuring that all datasets contribute similarly to the final score, independently of the number of image pairs in them.

The comprehensive results for fundamental matrix estimation are presented in Table~\ref{tab:fundamental_matrix_estimation}, with a visual summary of AUC@10$^\circ$ scores versus runtime also available in Fig.~\ref{fig:accuracy_curves_F}. Table~\ref{tab:fundamental_matrix_estimation} demonstrates the superior performance of SupeRANSAC.
With SuperPoint+LightGlue (SP+LG) correspondences, SupeRANSAC achieves the highest AUC scores across all reported thresholds (\eg, 0.59 at 10$^\circ$, compared to 0.53 for the next best LO-RSC from PoseLib, and 0.45 vs 0.39 at 5$^\circ$) and the lowest median pose error (1.85$^\circ$, significantly outperforming the 2.22$^\circ$ from LO-RSC PoseLib). While some RANSAC variants in OpenCV achieve faster raw execution times (0.01s), SupeRANSAC's runtime of 0.06s is highly competitive and is coupled with a vastly superior accuracy.

When utilizing denser RoMA correspondences, SupeRANSAC again leads by a substantial margin in all accuracy metrics, registering an AUC@10$^\circ$ of 0.72 (compared to ~0.65-0.66 for the closest competitors) and a median error of 1.09$^\circ$ (versus ~1.25-1.30$^\circ$ for others). The processing time for RoMA matches (0.81s for SupeRANSAC) is naturally higher due to the increased number of correspondences. 
Nevertheless, SupeRANSAC maintains efficiency relative to its accuracy achievements, proving faster than some other high-performing baselines.
In summary, across both sparse and dense feature types, SupeRANSAC consistently delivers substantial accuracy improvements over all baseline methods, establishing a new state-of-the-art in robust fundamental matrix estimation by offering an excellent accuracy-to-speed trade-off.

\subsection{Essential Matrix Estimation}

Essential matrix estimation is benchmarked on the same extensive set of datasets as fundamental matrix estimation. Detailed performance metrics are presented in Table~\ref{tab:essential_matrix_estimation}, with a visual summary of AUC@10$^\circ$ scores versus runtime also available in Fig.~\ref{fig:accuracy_curves_E}.

Table~\ref{tab:essential_matrix_estimation} highlights SupeRANSAC's strong performance in this task. With SuperPoint+LightGlue (SP+LG) correspondences, SupeRANSAC decisively leads in accuracy, achieving an AUC@10$^\circ$ of 0.66 and a med.\ error of 1.37$^\circ$. This represents a significant improvement over the next best performing methods, such as GC-RSC (AUC@10$^\circ$ of 0.60, med.\ error 1.72$^\circ$) and LO-RSC [Poselib] (AUC@10$^\circ$ of 0.59, med.\ error 1.70$^\circ$). SupeRANSAC's runtime of 0.08s for SP+LG features, while not the absolute fastest, offers an excellent accuracy-to-speed trade-off, being only marginally slower than some significantly less accurate OpenCV-based baselines.

When utilizing denser RoMA correspondences, SupeRANSAC again demonstrates its superiority, securing the top AUC scores across all thresholds (\eg, 0.74 at 10$^\circ$, compared to 0.68 for the closest competitors) and the lowest median error (1.04$^\circ$, with the next best at 1.19$^\circ$). At 0.62s with RoMA inputs, its runtime is competitive among high-accuracy methods and notably faster than several other baselines when processing these dense correspondences.

Fig.~\ref{fig:accuracy_curves_E} further corroborates these findings, positioning SupeRANSAC favorably in the accuracy-runtime spectrum. In summary, for essential matrix estimation, SupeRANSAC consistently provides substantial accuracy gains with both sparse and dense feature types, establishing itself as a state-of-the-art solution.
An interesting cross-comparison can also be drawn: SupeRANSAC, when configured for fundamental matrix estimation (as shown previously in Table~\ref{tab:fundamental_matrix_estimation}), often achieves pose accuracies that are comparable or even superior to those obtained by many of the baseline methods attempting direct essential matrix estimation (Table~\ref{tab:essential_matrix_estimation}). This underscores the capability of a highly robust fundamental matrix pipeline, like that embodied in SupeRANSAC, to yield excellent end-to-end pose information.

\begin{table}[t]
\centering
\scriptsize
\resizebox{1.0\columnwidth}{!}{
\begin{tabular}{c@{\hskip 8pt}l|ccccc@{\hskip 8pt}}
\toprule
     & Method & RR $\uparrow$ & avg RRE $\downarrow$ & avg RTE $\downarrow$ & med RRE $\downarrow$ & med RTE $\downarrow$ \\
\midrule
  \multirow{7}{*}{\rotatebox[origin=c]{90}{3DMatch}} & \textbf{SupeRANSAC} & 92.0 & \textbf{1.767} & \textbf{0.064} & \textbf{1.548} & \textbf{0.053} \\
   & GC-RSC [Author] & 92.4 & 1.787 & 0.066 & 1.555 & 0.054 \\
  & RANSAC [Open3D] & 91.5 & 2.031 & 0.070 & 1.735 & 0.057 \\
  & MAGSAC [Author] & 91.5 & 1.810 & 0.065 & 1.586 & \textbf{0.053} \\
  & MAGSAC++ [Author] & \textbf{92.5} & 1.876 & 0.067 & 1.614 & \textbf{0.053} \\
  & RANSAC [OpenCV] & 11.9 & 9.271 & 0.171 & 7.590 & 0.167 \\
  & w.\ Procrustes & 86.5 & 2.039 & 0.066 & 1.690 & 0.055 \\
  
\midrule
  \multirow{7}{*}{\rotatebox[origin=c]{90}{3DLoMatch}} & \textbf{SupeRANSAC} & 74.9 & \phantom{1}\textbf{2.851} & \textbf{0.090} & \textbf{2.441} & \textbf{0.075} \\
   & GC-RSC [Author] & 73.7 & \phantom{1}2.861 & 0.091 & 2.473 & 0.076 \\
  & RANSAC [Open3D] & 73.2 & \phantom{1}3.282 & 0.097 & 2.900 & 0.086 \\
  & MAGSAC [Author] & 74.9 & \phantom{1}2.927 & 0.092 & 2.607 & 0.079 \\
  & MAGSAC++ [Author] & \textbf{75.1} & \phantom{1}2.924 & 0.093 & 2.553 & 0.078 \\
  & RANSAC [OpenCV] & \phantom{1}4.9 & 13.163 & 0.228 & 9.799 & 0.224 \\
  & w.\ Procrustes & 58.7 & \phantom{1}3.915 & 0.107 & 3.026 & 0.087 \\
\bottomrule
\end{tabular}}
\caption{Performance of \textbf{rigid transformation estimation} on the 3DMatch~\cite{zeng20173dmatch} and 3DLoMatch~\cite{huang2021predator} datasets, utilizing matches generated by GeoTransformer~\cite{qin2023geotransformer}. The reported metrics include: registration recall (RR), calculated from the estimated inlier set of each method; as well as average and median Relative Rotation Errors (RRE) in degrees, and average and median Relative Translation Errors (RTE) in meters.}
\label{tab:rigid_pose}
\end{table}

\subsection{Homography Estimation}

Homography estimation is evaluated on the datasets previously described for relative pose, and additionally on the challenging HEB (Homography Estimation Benchmark) dataset~\cite{barath2023large}. HEB comprises 226,260 image pairs with ground truth homographies and approximately 4 million RootSIFT+MNN correspondences.

Performance on the general datasets using SP+LG and RoMA matches is detailed in Table~\ref{tab:homography_estimation}, with Fig.~\ref{fig:accuracy_curves_H} providing a visual summary of AUC@10$^\circ$ vs. runtime.
With SuperPoint+LightGlue (SP+LG) correspondences, Table~\ref{tab:homography_estimation} shows SupeRANSAC achieving the best accuracy across all AUC thresholds (\eg, 0.41 at 10$^\circ$, compared to ~0.39 for the nearest competitors like GC-RSC [Author] and MAGSAC++ [OpenCV]) and the lowest median pose error (3.36$^\circ$). While extremely fast, specialized methods like RHO [OpenCV] (effectively 0.000s) exist, SupeRANSAC (0.030s) delivers its leading accuracy with a still very competitive runtime.
Using denser RoMA correspondences, SupeRANSAC's superiority in accuracy is even more pronounced. It achieves an AUC@10$^\circ$ of 0.60 and a median error of 1.71$^\circ$, significantly ahead of other methods (next best AUC@10$^\circ$ around 0.59 for GC-RSC [Author]). Its runtime of 0.341s is reasonable for dense inputs, particularly considering the substantial accuracy gains, although methods like RHO [OpenCV] (0.005s) and VSAC (0.014s) remain faster but are less accurate.

On the HEB dataset (Table~\ref{tab:homography_heb}), evaluated with RootSIFT features, SupeRANSAC again demonstrates state-of-the-art accuracy. It achieves a mean Average Accuracy (mAA) of 0.51, substantially higher than the next best scores of 0.44 (from MAGSAC++ [Author] and LO-RSC [Poselib]). Furthermore, its median pixel error of 0.07 is exceptionally low, marking a clear improvement over all baselines (next best at 0.12). SupeRANSAC's runtime (0.027s) is competitive within this benchmark. It is important to reiterate that while SupeRANSAC does not always yield the highest raw inlier count (\eg, LMEDS reports more in Table~\ref{tab:homography_heb}), this metric is less indicative of true geometric accuracy, particularly when comparing diverse robust estimators with varying scoring functions or thresholding mechanisms. The mAA scores versus runtime for the HEB dataset are also visualized in Fig.~\ref{fig:homography_heb_curve}.

Overall, for homography estimation, SupeRANSAC consistently delivers the highest accuracy across diverse datasets, feature types, and evaluation metrics, establishing a new benchmark for robust homography computation.

\subsection{Absolute Pose Estimation}

We assess absolute pose estimation performance by integrating SupeRANSAC into the HLoc visual localization pipeline~\cite{sarlin2019coarse} and evaluating on the challenging Aachen Day-Night~\cite{sattler2018benchmarking} and InLoc~\cite{taira2018inloc} benchmarks. The detailed results, reporting Area Under the Curve (AUC) scores at three precision thresholds for each dataset subset, are presented in Table~\ref{tab:absolute_pose}. The thresholds for Aachen Day-Night are (0.25m, 2$^\circ$), (0.5m, 5$^\circ$), and (5m, 10$^\circ$), while for InLoc they are (0.25m, 10$^\circ$), (0.5m, 10$^\circ$), and (1m, 10$^\circ$).

On the Aachen Day-Night dataset, SupeRANSAC's performance varies between conditions. For the Day subset, it achieves highly competitive results (AUCs of 80.7 / 93.6 / 97.8). While LO-RANSAC implementations from COLMAP and PoseLib show slightly higher accuracy at the strictest (0.25m, 2$^\circ$) threshold (88.5 and 88.1 respectively), SupeRANSAC matches their top performance at the (5m, 10$^\circ$) threshold. In contrast, on the more challenging Night subset, SupeRANSAC demonstrates a clear advantage, achieving the best AUC scores across the stricter and medium thresholds (78.5 / 89.5 / 97.4). This notably surpasses the next best methods, LO-RANSAC [COLMAP] (70.2 / 88.0 / 97.4) and LO-RANSAC [PoseLib] (70.7 / 87.4 / 97.4), particularly at the (0.25m, 2$^\circ$) threshold.

For the indoor InLoc dataset, SupeRANSAC consistently delivers strong results. On the DUC1 subset, it achieves the highest AUC scores across all three evaluation thresholds (46.0 / 67.2 / 77.3), outperforming all baselines including LO-RANSAC [PoseLib] (44.4 / 66.2 / 76.3). On the DUC2 subset, SupeRANSAC leads at the strictest threshold (AUC of 53.4 vs. 50.4 for LO-RANSAC [PoseLib]) and matches the top performance of LO-RANSAC [PoseLib] at the medium threshold (both at 70.2), while all top methods converge at the loosest threshold (73.3).

Overall, SupeRANSAC exhibits highly competitive and often superior performance for absolute pose estimation. It particularly excels in challenging scenarios, such as nighttime localization on Aachen Day-Night and across the varied conditions of the InLoc benchmark, establishing itself as a state-of-the-art robust estimator for this task.

\subsection{Rigid Transformation}

Rigid transformation estimation from 3D point clouds is evaluated on the 3DMatch~\cite{zeng20173dmatch} and the more 3DLoMatch~\cite{huang2021predator} datasets, utilizing pre-computed correspondences from GeoTransformer~\cite{qin2023geotransformer}. The detailed performance metrics are presented in Table~\ref{tab:rigid_pose}, including Registration Recall (RR), and average and median Relative Rotation Errors (RRE) in degrees, and Relative Translation Errors (RTE) in meters.

On the 3DMatch dataset, SupeRANSAC demonstrates high precision. It achieves the best average RRE (1.767$^\circ$), average RTE (0.064m), and median RRE (1.548$^\circ$). Furthermore, it ties for the best median RTE (0.053m) with MAGSAC and MAGSAC++. Its Registration Recall (RR) of 92.0\% is highly competitive, closely approaching the top score of 92.5\% achieved by MAGSAC++. In contrast, standard RANSAC from OpenCV shows significantly poorer performance across all metrics on this dataset.

This trend of superior precision continues on the 3DLoMatch dataset, which features scenes with lower overlap. Here, SupeRANSAC consistently outperforms all baselines across all reported error metrics. It achieves the lowest average RRE (2.851$^\circ$), average RTE (0.090m), median RRE (2.441$^\circ$), and median RTE (0.075m). Its RR of 74.9\% is again very competitive, nearly matching the top RR of 75.1\% from MAGSAC++ and clearly outperforming most other methods.

In summary, for rigid transformation estimation, while SupeRANSAC's Registration Recall is on par with other leading methods like MAGSAC++, it consistently delivers the most accurate transformations in terms of both rotation and translation errors (both average and median metrics) across standard and low-overlap scenarios. Let us note that RR can be influenced by inlier definitions. 
Thus, SupeRANSAC's consistently lower geometric errors provide a strong testament to its robustness and precision for 3D point cloud registration.

\begin{figure*}[t!]
    \centering
    \begin{subfigure}[t]{1.0\textwidth}
        \centering
        \includegraphics[width=0.32\textwidth]{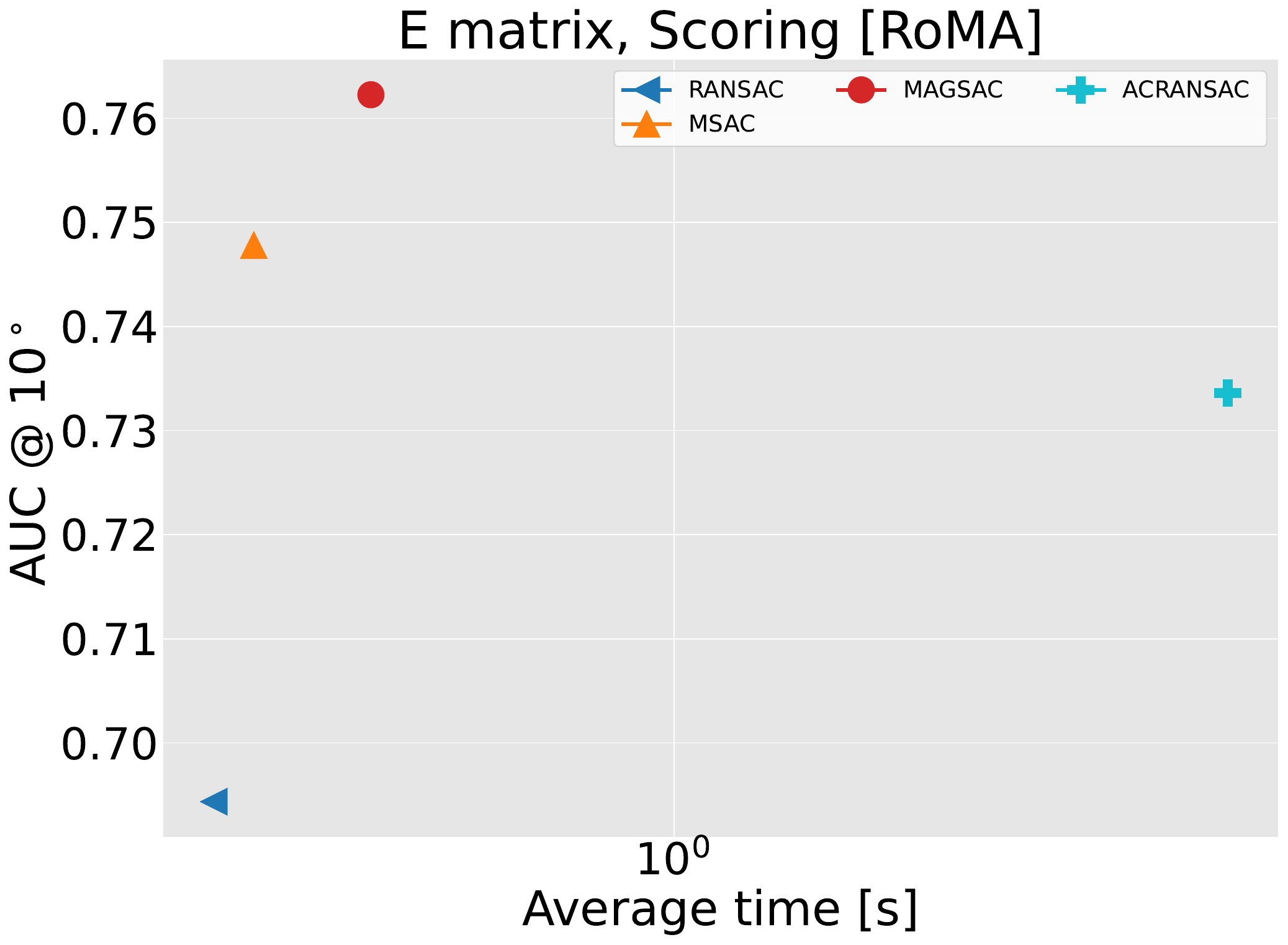}
        \includegraphics[width=0.32\textwidth]{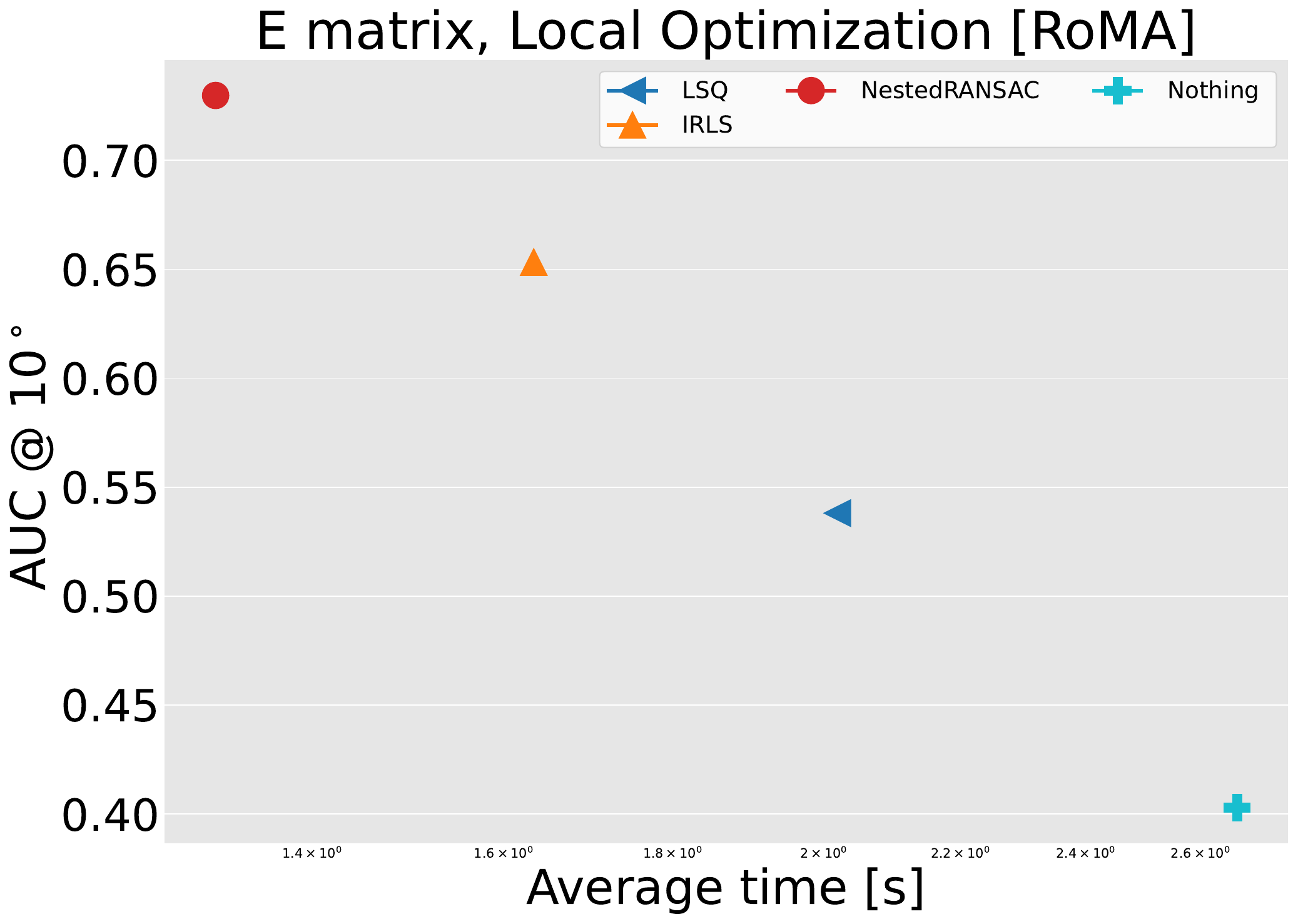}
        \includegraphics[width=0.32\textwidth]{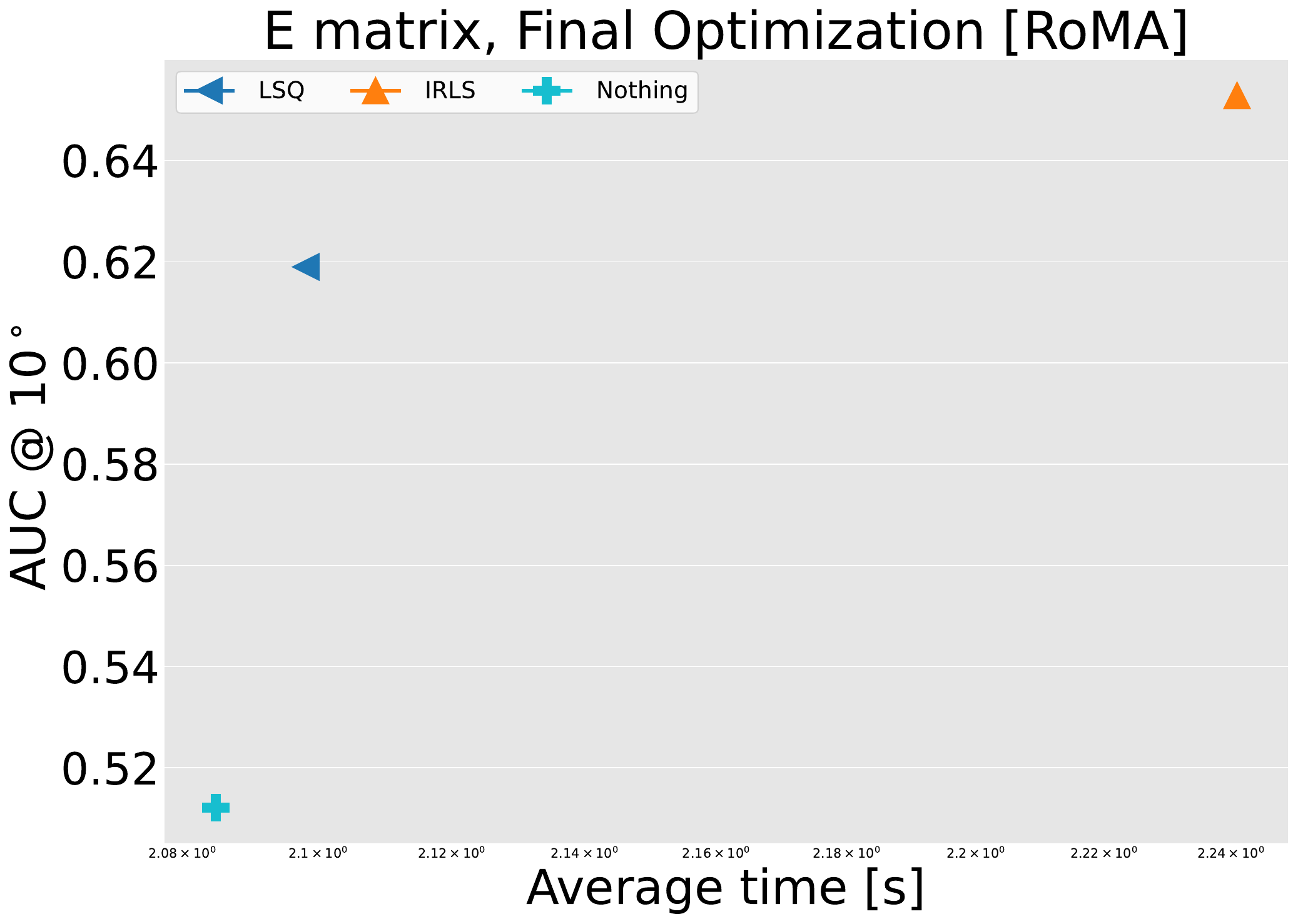}
        \caption{Essential matrix}
    \end{subfigure}
    \begin{subfigure}[t]{1.0\textwidth}
        \centering
        \includegraphics[width=0.32\textwidth]{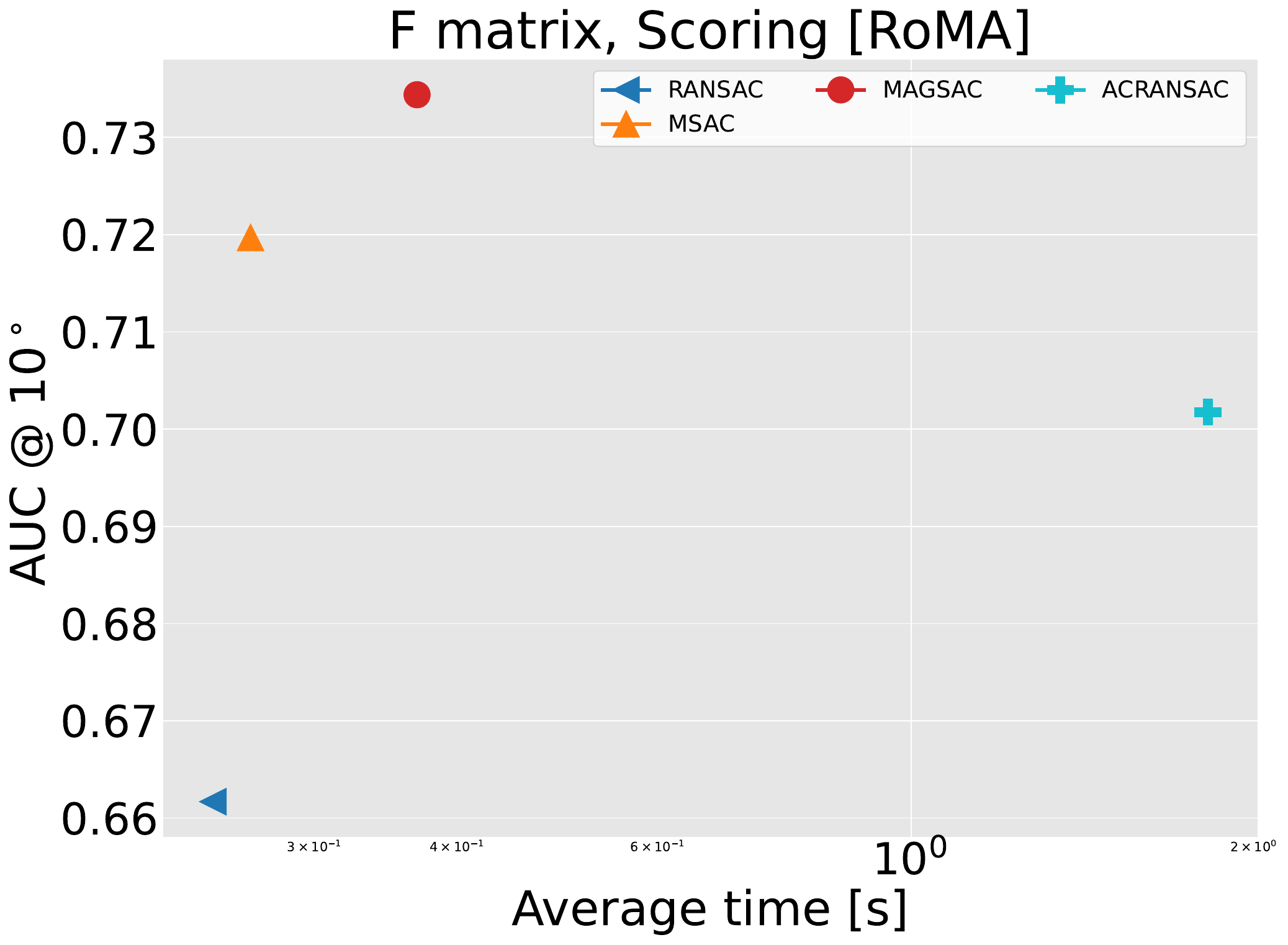}
        \includegraphics[width=0.32\textwidth]{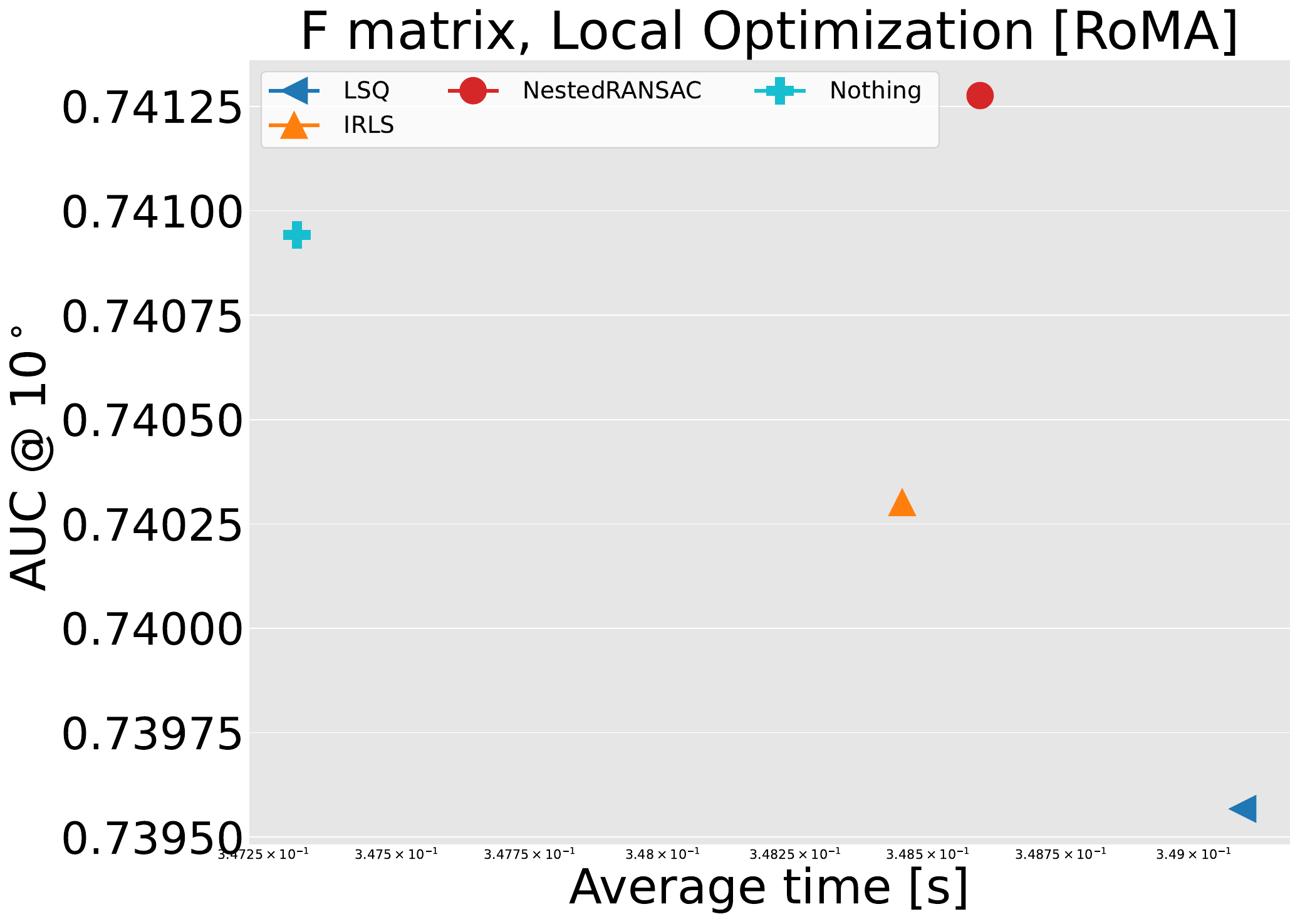}
        \includegraphics[width=0.32\textwidth]{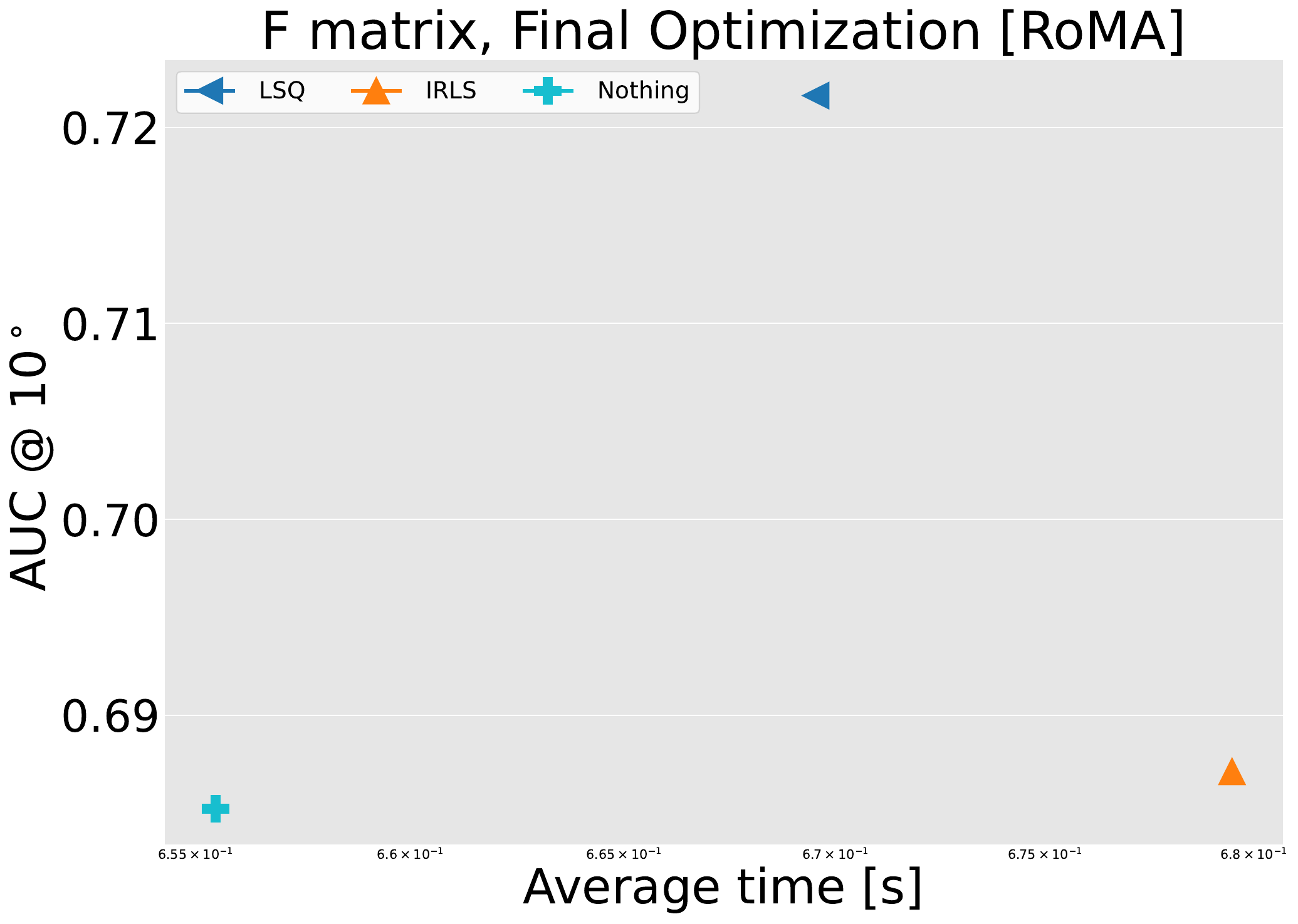}
        \caption{Fundamental matrix}
    \end{subfigure}
    \begin{subfigure}[t]{1.0\textwidth}
        \centering
        \includegraphics[width=0.32\textwidth]{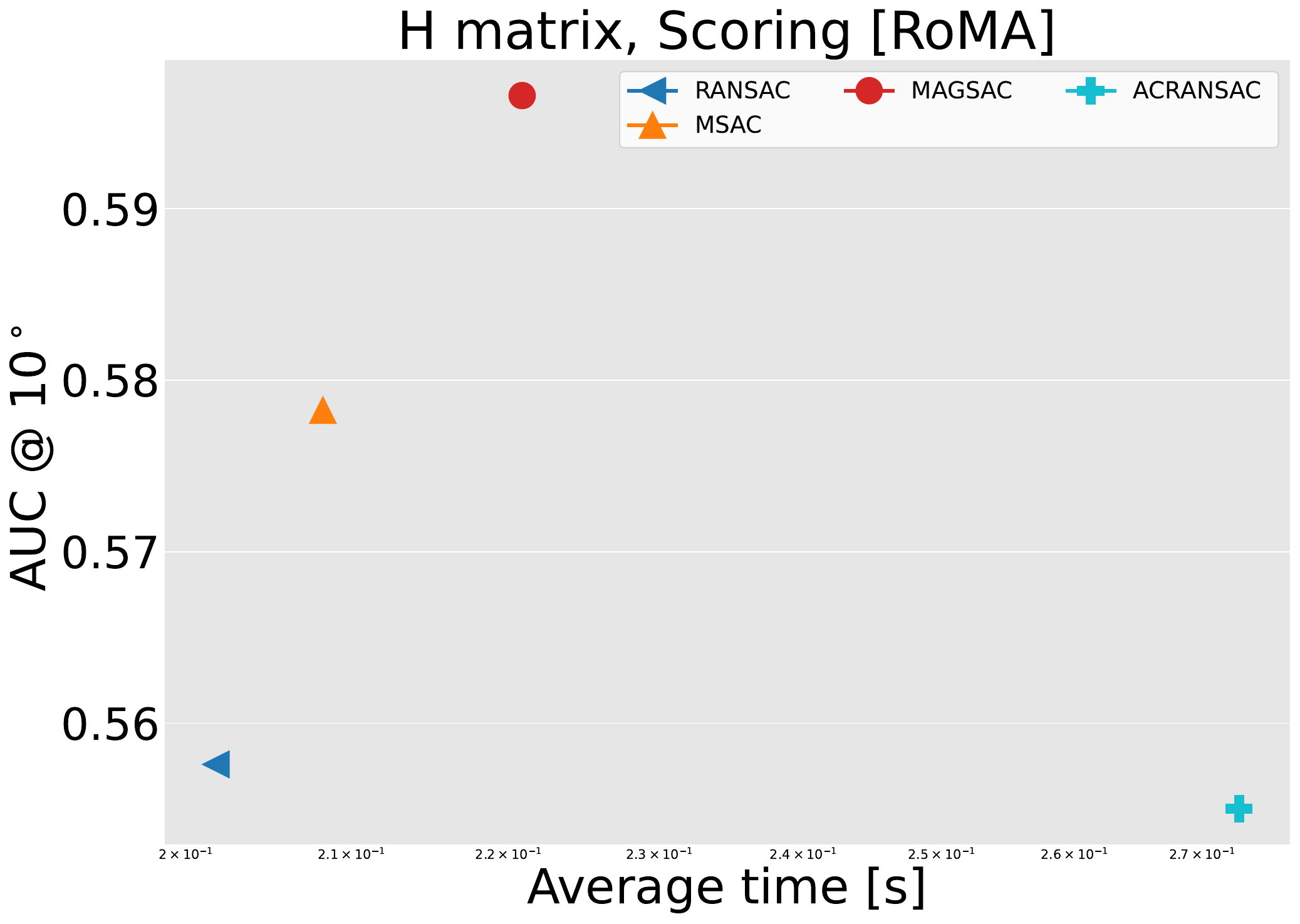}
        \includegraphics[width=0.32\textwidth]{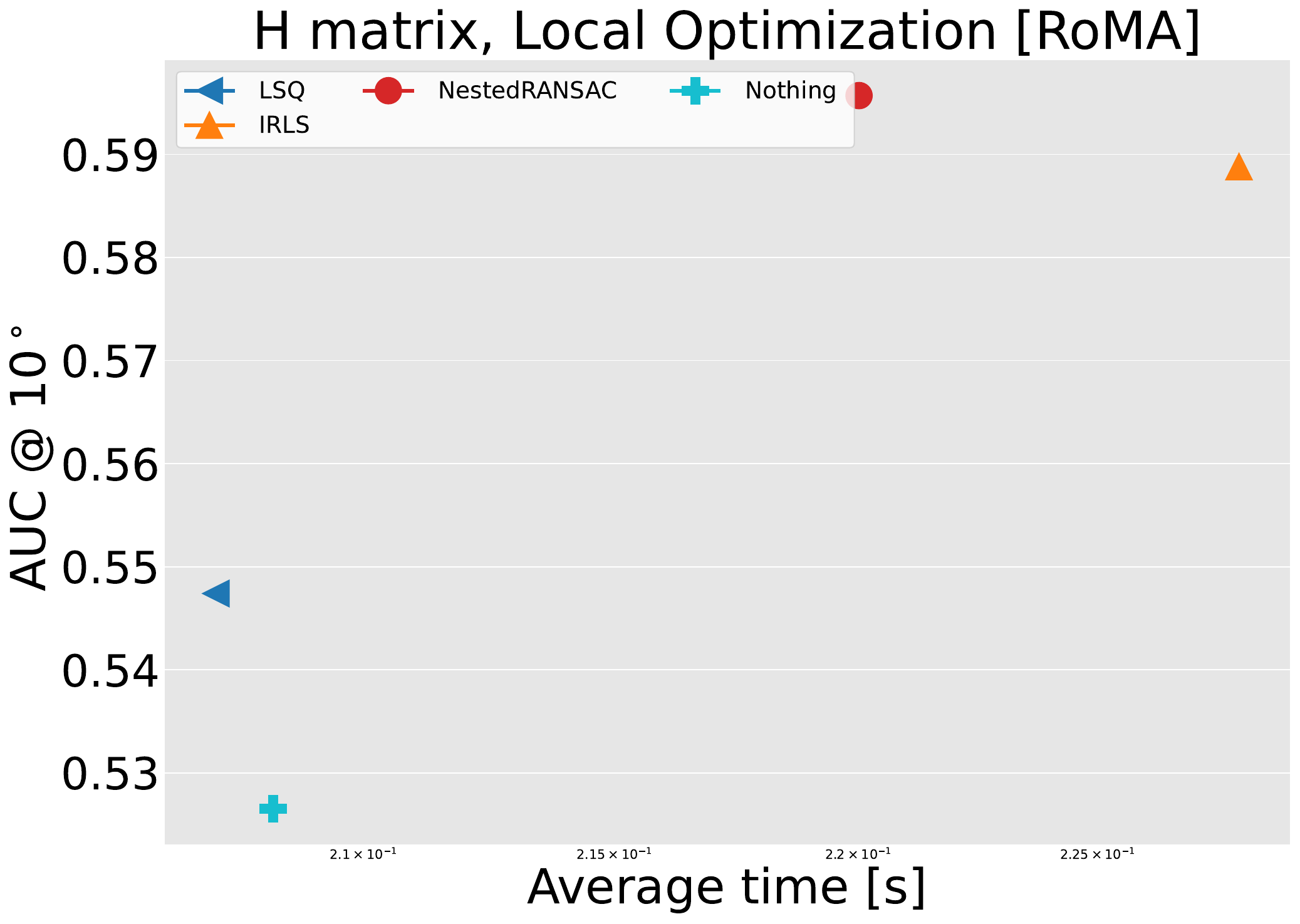}
        \includegraphics[width=0.32\textwidth]{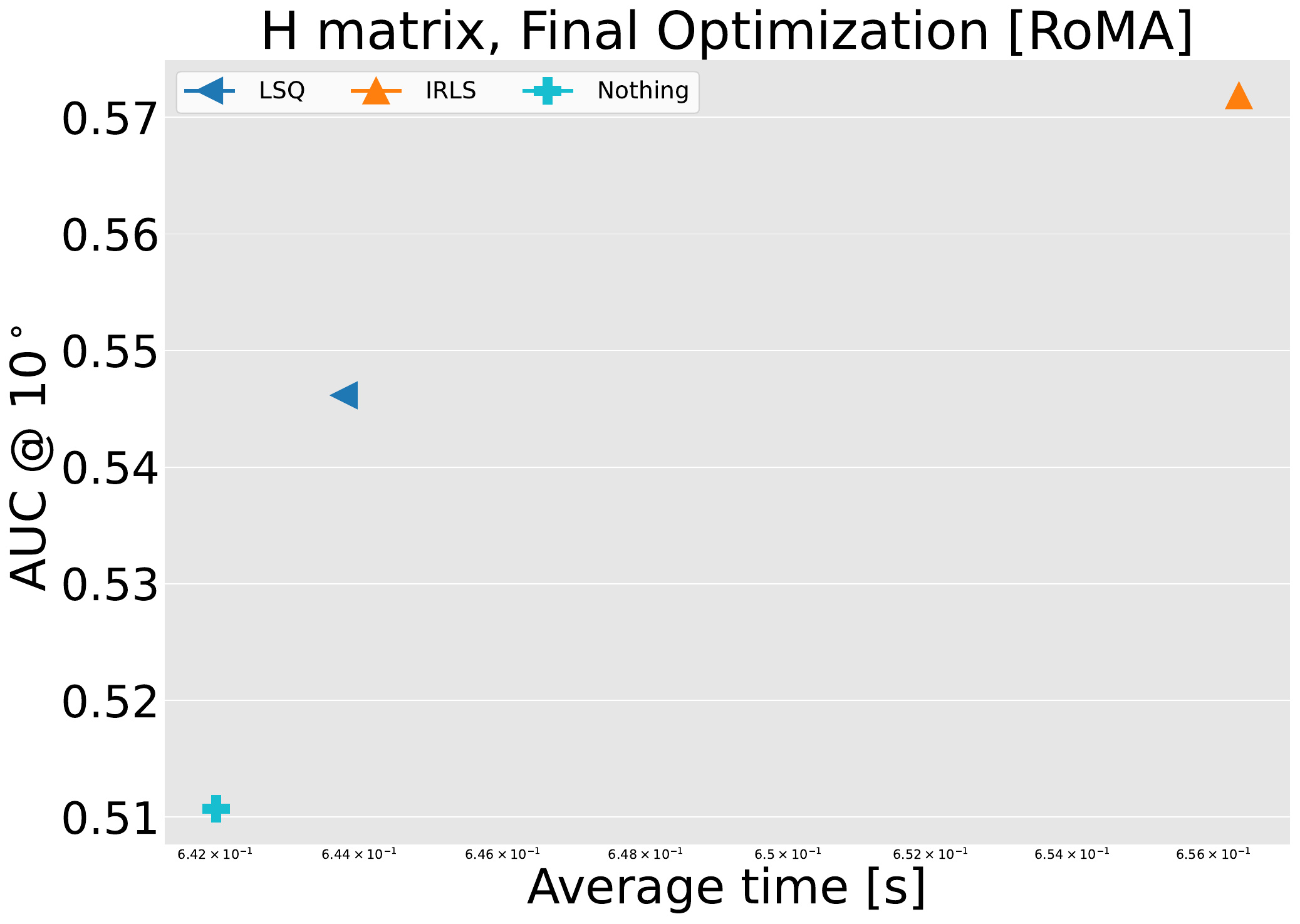}
        \caption{Homography}
    \end{subfigure}
    \caption{\textbf{Ablation studies of SupeRANSAC components.} This figure illustrates the performance impact when varying key algorithmic choices. For each geometric problem shown -- (a) Essential matrix, (b) Fundamental matrix, and (c) Homography (corresponding to image rows from top to bottom) -- the three plots (from left to right) demonstrate the effects of ablating the scoring technique, the local optimization (LO) strategy, and the final optimization (FO) strategy, respectively. All evaluations were performed on the 1200-pair tuning set.}
    \label{fig:ablations}
\end{figure*}

\subsection{Ablation Studies}

To validate our design choices and systematically quantify the impact of key algorithmic components within the SupeRANSAC framework, we conducted a comprehensive set of ablation studies. These experiments focused on evaluating alternative strategies for model scoring, local optimization (LO), and final optimization (FO). The performance variations are summarized in Figure~\ref{fig:ablations}, using the same consolidated tuning set of 1200 image pairs (200 from each of the six datasets) employed for the main parameter tuning, ensuring consistency in our evaluation. 

\paragraph{Impact of Scoring Function}

We first investigated the choice of the model scoring function, comparing SupeRANSAC's default MAGSAC++~\cite{barath2020magsac++} against other prominent methods such as standard RANSAC inlier counting, MSAC~\cite{torr2000mlesac}, and a contrario RANSAC as detailed in the left column of Figure~\ref{fig:ablations}. 
The results consistently demonstrate that MAGSAC++ yields the highest performance across all evaluated geometric problems. This underscores the benefits of its marginalization strategy, which provides robustness against the strict thresholding of traditional methods and better handles varying noise levels, leading to more accurate model selection.

\paragraph{Efficacy of Local Optimization Strategies}

In the second column of Figure~\ref{fig:ablations}, the role of local optimization was examined by comparing different LO techniques. SupeRANSAC's default approach utilizes GC-RANSAC~\cite{barath2018graph} (noted as NestedRANSAC in the figures). Our ablations assessed strategies such as Nested RANSAC, least squares fitting on the inliers, iteratively reweighted least squares fitting, and no LO. The findings indicate that employing a sophisticated LO like Nested RANSAC leads to the highest accuracy scores for most problems. For fundamental matrix estimation, while advanced LO still provides benefits, the improvement over simpler LO techniques was observed to be less pronounced compared to its impact on other problems like essential matrix or homography estimation. This suggests that while crucial, the specific form of advanced LO for F-matrix might offer slightly more flexibility.

\paragraph{Choice of Final Optimization}

Finally, we ablated the final optimization (FO) stage, which refines the best model found by the main RANSAC loop. 
For essential matrix and homography estimation, an iteratively reweighted least-squares (IRLS) fitting, specifically using robust Cauchy weights, proved to be the most effective FO strategy, yielding noticeable gains in precision.
An interesting finding emerged for fundamental matrix estimation: a simpler approach, effectively performing a single non-minimal model refinement, yielded the best results. 
It is crucial to note that our non-minimal solver for fundamental matrices (as described in Section~\ref{sec:nonminimal_estimation}) already incorporates a Levenberg-Marquardt optimization that enforces geometric constraints (\eg, rank-2). The ablation results suggest that this integrated LM refinement is sufficiently powerful, and further applying a separate, complex IRLS stage for FO does not offer additional benefits and may even slightly dilute the precision already achieved by the specialized non-minimal solver.

\begin{figure*}[t!]
    \centering
    \begin{subfigure}[t]{0.32\textwidth}
        \centering
        \includegraphics[width=1.0\columnwidth]{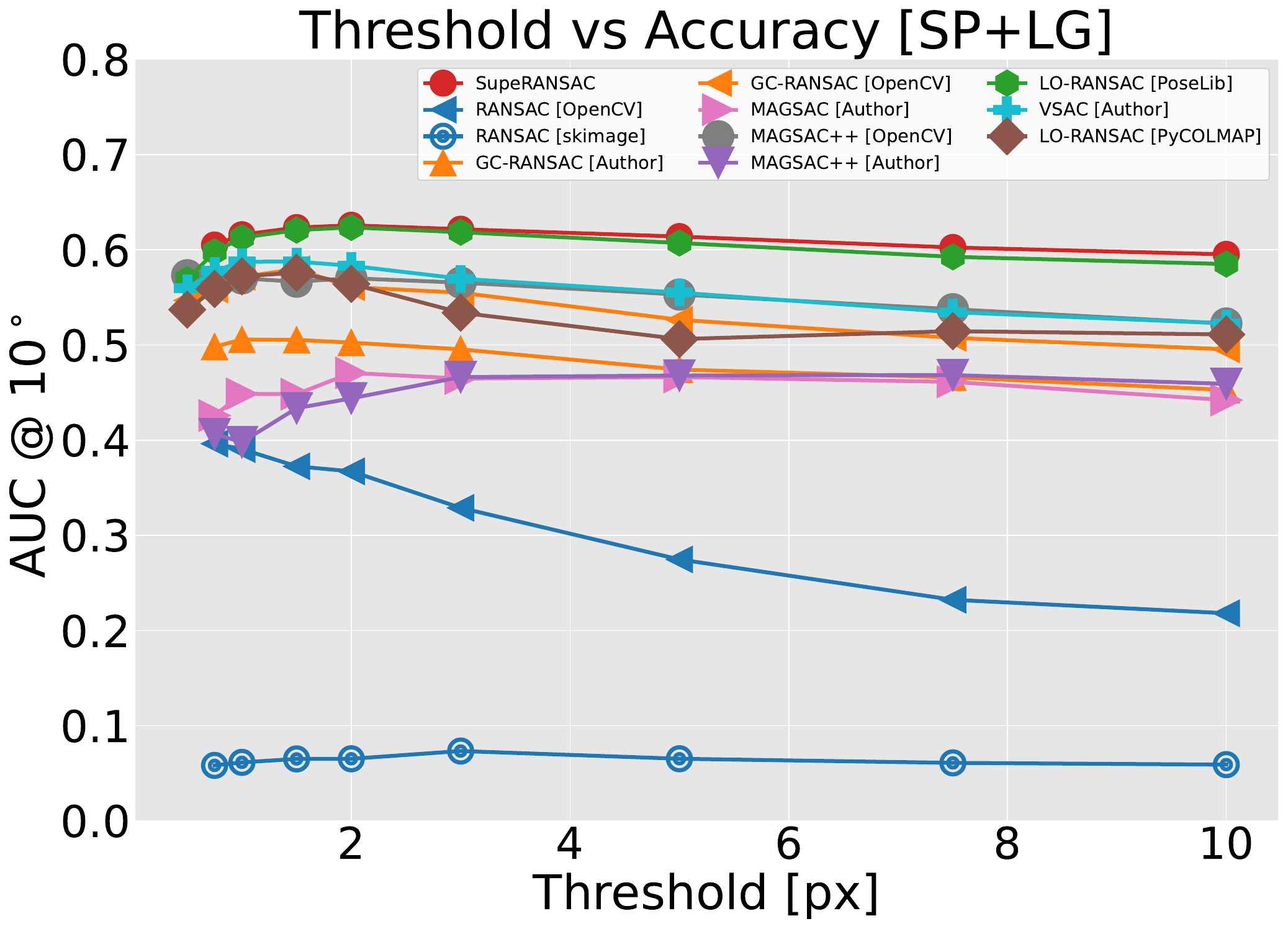}\\
        \includegraphics[width=1.0\columnwidth]{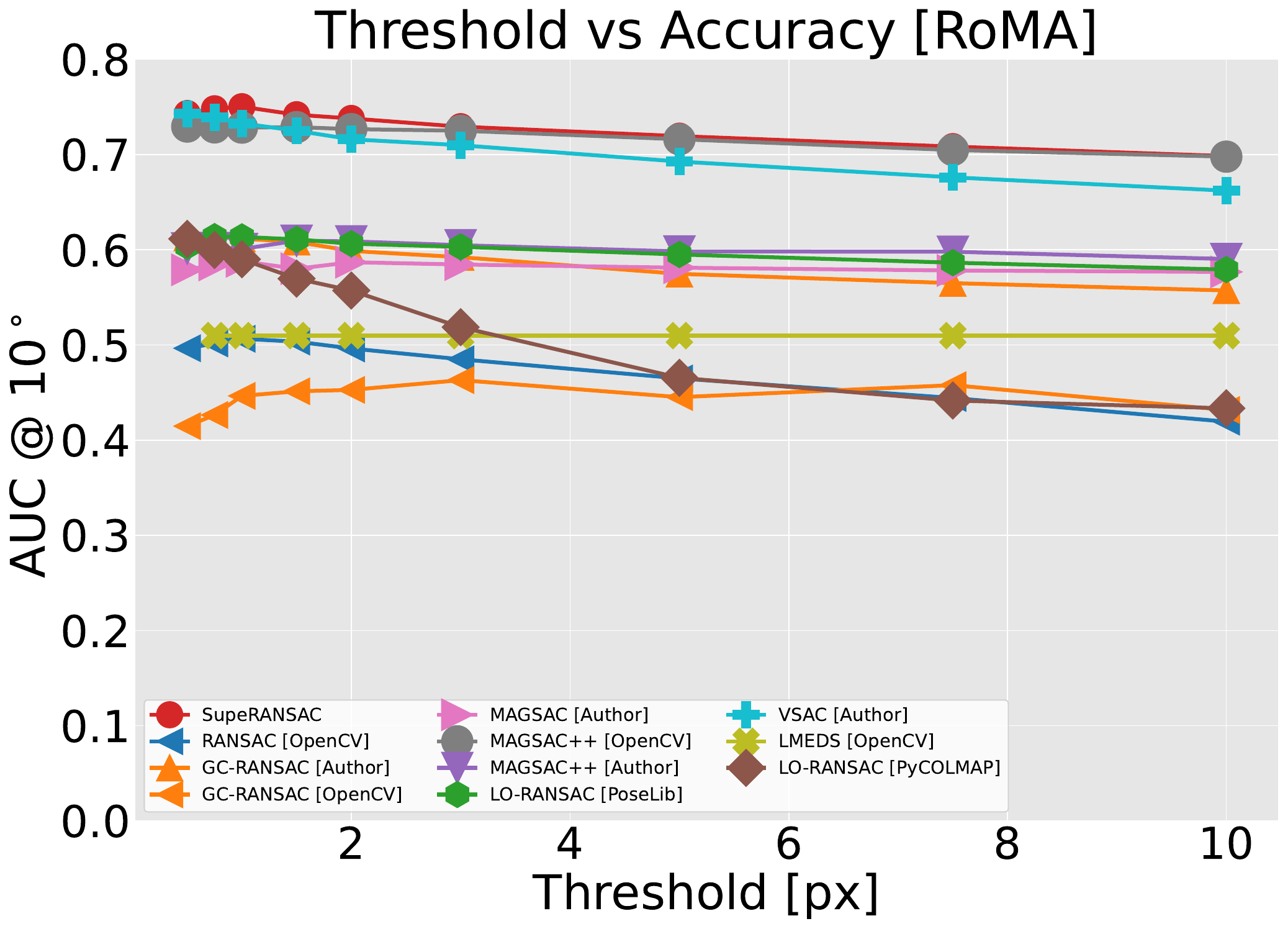}
        \caption{Fundamental matrix}
        \label{fig:tuning_F}
    \end{subfigure}%
    \begin{subfigure}[t]{0.325\textwidth}
        \centering
        \includegraphics[width=1.0\columnwidth]{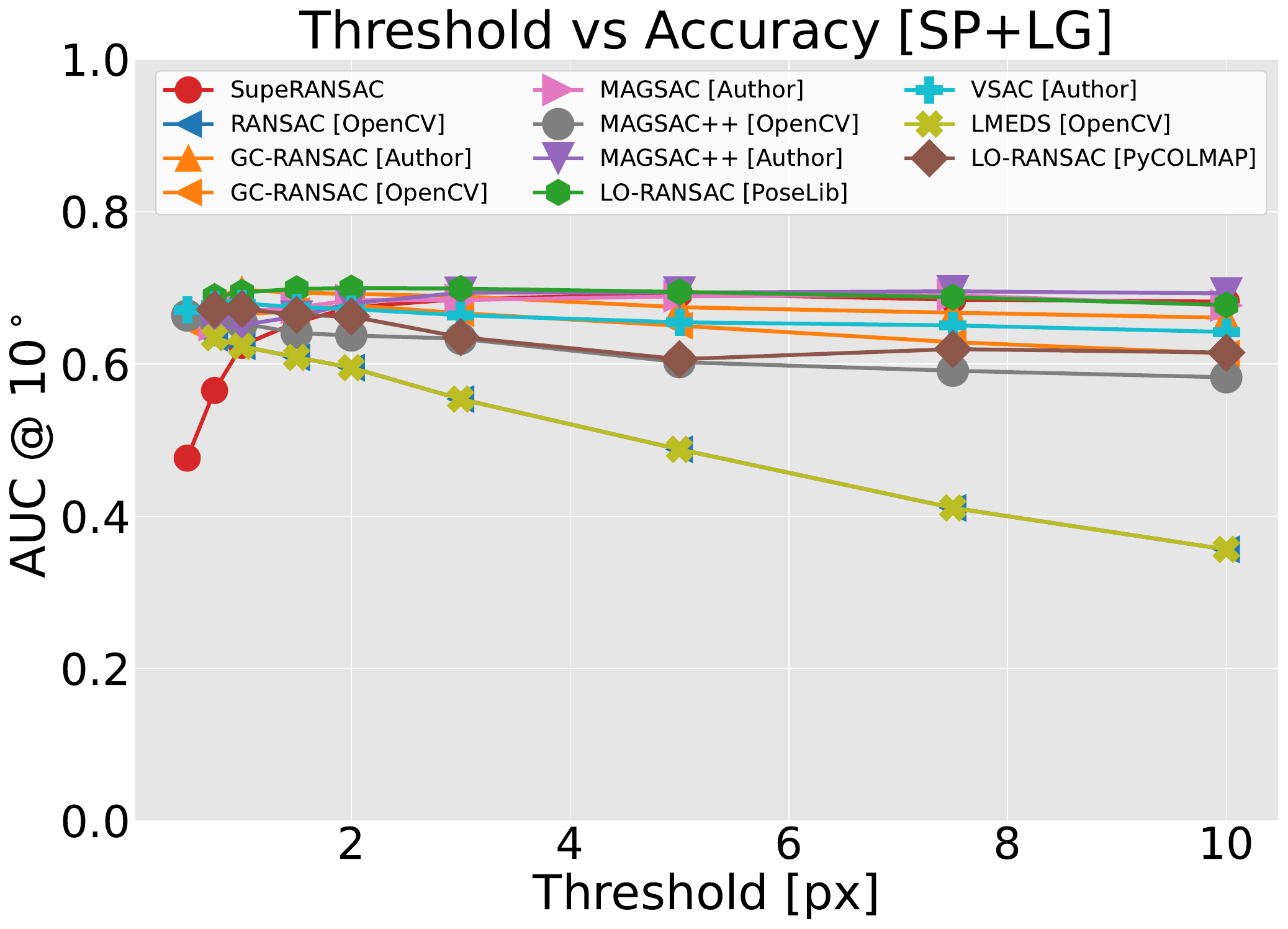}\\
        \includegraphics[width=1.0\columnwidth]{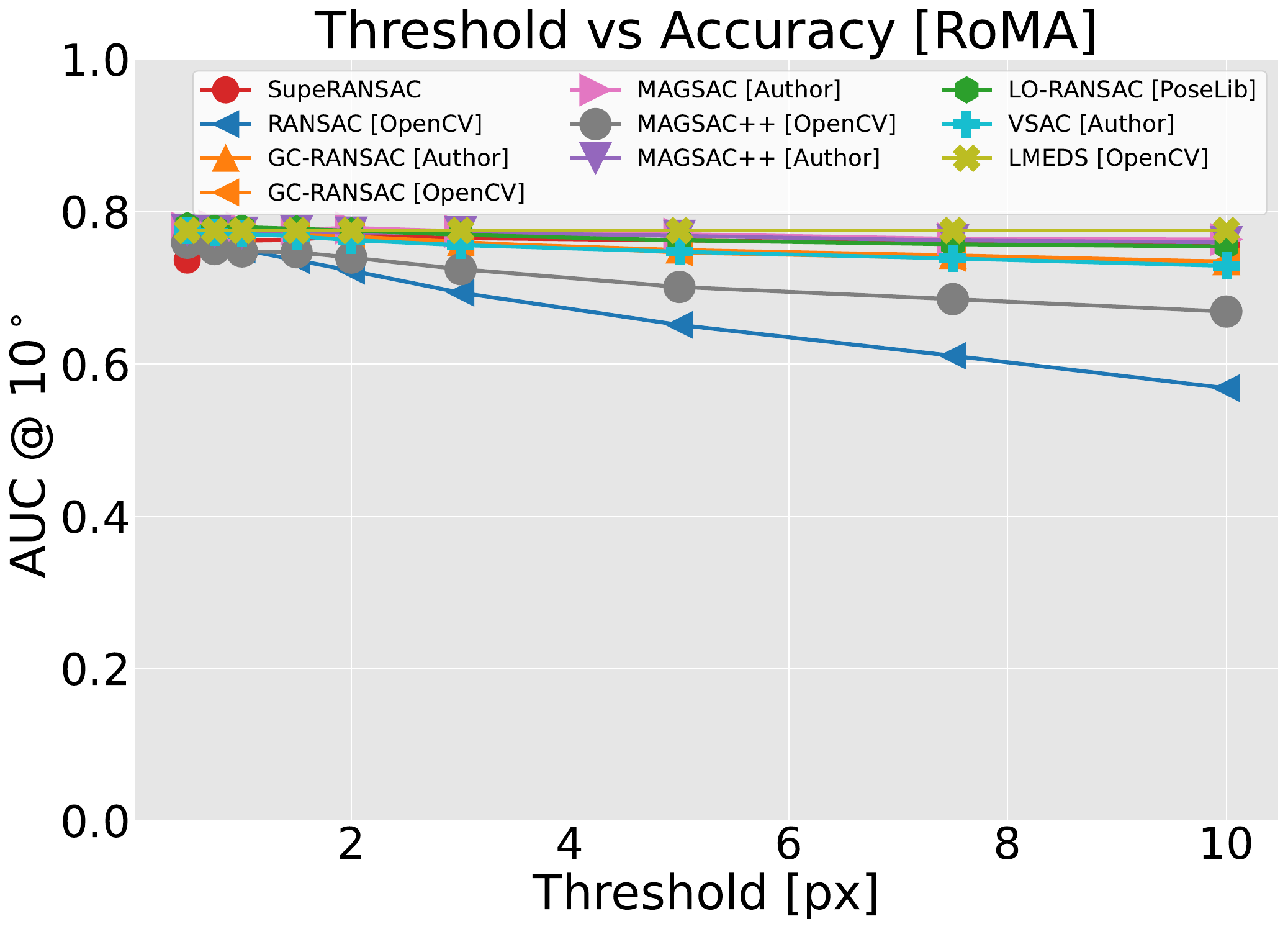}
        \caption{Essential matrix}
        \label{fig:tuning_E}
    \end{subfigure}
    \begin{subfigure}[t]{0.325\textwidth}
        \centering
        \includegraphics[width=1.0\columnwidth]{figures/homography_matrix_splg.pdf}\\
        \includegraphics[width=1.0\columnwidth]{figures/homography_matrix_roma.pdf}
        \caption{Homography}
        \label{fig:tuning_H}
    \end{subfigure}
    \caption{\textbf{Inlier-outlier threshold tuning for robust estimators} across (a) fundamental matrix, (b) essential matrix, and (c) homography estimation. These plots illustrate Area Under the Curve (AUC@10$^\circ$) scores as a function of the inlier-outlier threshold value (in pixels). Results are averaged over six datasets (ScanNet1500~\cite{dai2017scannet,sarlin2020superglue}, PhotoTourism~\cite{snavely2006photo,jin2021image}, LaMAR~\cite{sarlin2022lamar}, 7Scenes~\cite{glocker2013real}, ETH3D~\cite{schops2019bad}, and KITTI~\cite{geiger2013vision}), with each curve generated using 200 randomly selected image pairs from each respective dataset. The evaluation distinguishes between two feature types: SuperPoint~\cite{detone2018superpoint}+LightGlue~\cite{lindenberger2023lightglue} (top row) and RoMA~\cite{edstedt2024roma} matches (bottom). In these plots, higher AUC values (Y-axis) indicate superior accuracy; robust performance is characterized by maintaining high AUC scores across a broad range of threshold values (X-axis).}
    \label{fig:accuracy_curves}
\end{figure*}

\subsection{Threshold Tuning}

The selection of an appropriate inlier-outlier threshold is critical for the performance of any RANSAC-based robust estimator. To analyze the sensitivity of SupeRANSAC and baseline methods to this parameter, we conducted a series of experiments, plotting their performance as a function of varying inlier-outlier thresholds. 
Figures~\ref{fig:tuning_F}, \ref{fig:tuning_E}, and \ref{fig:tuning_H} illustrate these sensitivity curves, showcasing AUC@10$^\circ$ scores for fundamental matrix, essential matrix, and homography estimation, respectively. These evaluations were performed on the consolidated tuning set of 1200 image pairs (200 randomly selected pairs from each of the six datasets), identical to the set used for parameter selection in our main experimental validation.

A key observation from these figures is that the proposed SupeRANSAC consistently ranks at or near the top in terms of peak accuracy across a broad spectrum of inlier-outlier threshold values. This behavior underscores SupeRANSAC's robustness to the precise choice of this hyperparameter. While many robust estimators exhibit a narrow optimal range for the threshold, achieving peak performance only within a small window and degrading significantly otherwise, SupeRANSAC demonstrates a more stable performance profile. For instance, even if the chosen threshold is not perfectly optimal for a given scene, SupeRANSAC tends to maintain a higher level of accuracy compared to methods that are more acutely sensitive.

Furthermore, the figures reveal that SupeRANSAC not only exhibits this robustness but also generally achieves the highest peak performance on this tuning set when compared to the other methods at their respective optimal thresholds within these plots. While the margin over the next best method can vary depending on the specific problem and threshold, SupeRANSAC's ability to deliver leading accuracy over a wider range of thresholds is a significant practical advantage, simplifying the tuning process and enhancing its reliability across diverse scenarios. This characteristic, combined with its overall superior performance as demonstrated in the main experimental sections (using the single threshold selected via this tuning process), solidifies its position as a highly effective and dependable robust estimation framework.

\section{Conclusions}

We propose a new robust estimation framework that improves upon baselines across popular vision problems on public large-scale datasets by a significant margin. 
SupeRANSAC demonstrates that robust estimation relies more on the "bells and whistles" of the framework than on individual algorithmic improvements. 
However, our work also highlights important insights  --  \eg, the significant role of local and final optimization, showing that running a few iterations of Levenberg-Marquardt optimization is practical and greatly enhances performance, compared with baselines. 
We believe that this framework will be beneficial to the community.

\ifCLASSOPTIONcaptionsoff
  \newpage
\fi



\bibliographystyle{IEEEtran}
\bibliography{main}

\end{document}